\documentclass{article}

\usepackage{microtype}
\usepackage{graphicx}
\usepackage{booktabs} 

\usepackage{hyperref}

\usepackage[accepted]{icml2025}

\usepackage{amsmath}
\usepackage{amssymb}
\usepackage{mathtools}
\usepackage{amsthm}

\usepackage[capitalize,noabbrev]{cleveref}

\theoremstyle{plain}

\theoremstyle{definition}

\theoremstyle{remark}

\usepackage[textsize=tiny]{todonotes}

\usepackage{booktabs}
\usepackage{tabularx}
\usepackage{amsmath}
\usepackage{amssymb}  
\usepackage{xcolor}
\usepackage{multirow}
\usepackage{hyperref}
\usepackage{float}

\icmltitlerunning{GraphSOS: Graph Sampling and Order Selection to Help LLMs Understand Graphs Better}

\begin{document}

\twocolumn[
\icmltitle{GraphSOS: Graph Sampling and Order Selection to Help LLMs Understand Graphs Better}



\icmlsetsymbol{equal}{*}

\begin{icmlauthorlist}
\icmlauthor{Xu Chu}{equal,pku}
\icmlauthor{Hanlin Xue}{equal,pku}
\icmlauthor{Zhijie Tan}{pku}
\icmlauthor{Bingce Wang}{pku}
\icmlauthor{Tong Mo}{pku}
\icmlauthor{Weiping Li}{pku}
\end{icmlauthorlist}

\icmlaffiliation{pku}{School of Software and Microelectronics, Peking University, Beijing, China}

\icmlcorrespondingauthor{Xu Chu}{chuxu@stu.pku.edu.cn}
\icmlcorrespondingauthor{Weiping Li}{wpli@ss.pku.edu.cn}
\icmlkeywords{Machine Learning, ICML}

\vskip 0.3in
]



\makeatletter\def\Hy@Warning#1{}\makeatother
\printAffiliationsAndNotice{\icmlEqualContribution} 

\begin{abstract}
The success of Large Language Models (LLMs) in various domains has led researchers to apply them to graph-related problems by converting graph data into natural language text. However, unlike graph data, natural language inherently has sequential order. We observe a counter-intuitive fact that when the order of nodes or edges in the natural language description of a graph is shuffled, despite describing the same graph, model performance fluctuates between high performance and random guessing. Additionally, due to LLMs' limited input context length, current methods typically randomly sample neighbors of target nodes as representatives of their neighborhood, which may not always be effective for accurate reasoning. To address these gaps, we introduce GraphSOS (Graph \underline{S}ampling and \underline{O}rder \underline{S}election). This novel model framework features an Order Selector Module to ensure proper serialization order of the graph and a Subgraph Sampling Module to sample subgraphs with better structure for better reasoning. Furthermore, we propose Graph CoT obtained through distillation, and enhance LLM's reasoning and zero-shot learning capabilities for graph tasks through instruction tuning. Experiments on multiple datasets for node classification and graph question-answering demonstrate that GraphSOS improves LLMs' performance and generalization ability on graph tasks.
\end{abstract}

\section{Introduction}
The recent success of Large Language Models (LLMs)~\cite{Touvron2023LLaMAOA,Bai2023QwenTR} motivates researchers to explore their potential in handling tasks across various modalities, including vision, speech, and tabular data~\cite{Li2023BLIP2BL,Fang2024LLaMAOmniSS,Xia2024ChartXC}, as well as graph data. As a non-Euclidean geometric structure, graphs are indispensable in representing and solving numerous applications, including social network analysis~\cite{kumar2022influence,liu2024rumor}, recommendation systems~\cite{fan2019graph,luo2024collaborative}, and spatiotemporal prediction~\cite{pareja2020evolvegcn,zhu2023wingnn}. 
Many Graph LLM studiess~\cite{Fatemi2023TalkLA,Guo2023GPT4GraphCL,Tang2023GraphGPTGI,Chen2024GraphWizAI} focus on converting graph data into natural language text and inputting it along with questions into closed-source LLMs or LLMs fine-tuned with graph tasks. LLMs complete graph tasks based on their inherent knowledge and reasoning capabilities, such as node classification~\cite{Tang2023GraphGPTGI} and graph question-answering~\cite{Chen2024GraphWizAI}.

\begin{figure}[t]
\centering
\includegraphics[width=0.49\textwidth]{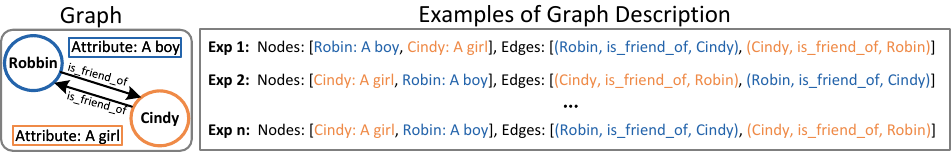} 
\caption{Converting a graph into natural language description. Elements in both node and edge lists can be arranged in any order to represent the same graph.}
\label{pic: intro}
\end{figure}

\begin{figure*}[t]
\centering
\includegraphics[width=0.98\textwidth]{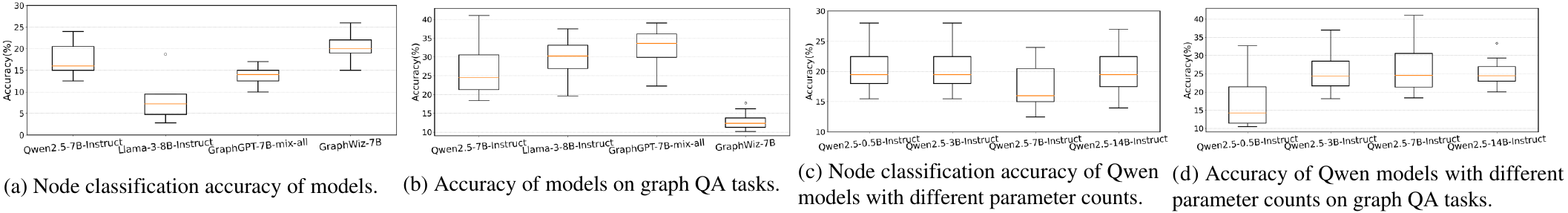} 
\caption{Zero-shot performance of models with different orders of node and edge.}
\label{fig:boxplot}
\end{figure*}

Despite promising results, we identify two concerning issues and raise questions accordingly. \textbf{Question I}: Do LLMs truly understand and process graph topological structures correctly? In graph learning using LLM as predictor~\cite{Chen2023ExploringTP}, the mainstream paradigm serializes graphs using natural language descriptions of nodes and edges~\cite{Ren2024ASO}, as shown in Figure~\ref{pic: intro}. Since natural language sequences are one-dimensional, the same graph can have multiple equivalent descriptive orders. However, research on both LLMs and Multimodal Large Language Models (MLLMs) indicates that LLMs/MLLMs have better prompt orders (though no universal optimal order exists for models or tasks)~\cite{Lu2021FantasticallyOP,tan2024order}. Models perform well when the ordering is correct, while other orders lead to near-random performance, which may affect model performance on graph tasks. We conduct node classification tasks on the text-attributed graph dataset Cora ~\cite{Chen2023ExploringTP} and graph question-answering tasks on the movie knowledge graph dataset MetaQA ~\cite{Zhang2017VariationalRF}. We test LLMs of different architectures and scales (Qwen 2.5~\cite{qwen2_5}, LLaMA 3~\cite{Dubey2024TheL3}) and two advanced Graph LLMs (GraphGPT~\cite{Tang2023GraphGPTGI}, GraphWiz~\cite{Chen2024GraphWizAI}) under zero-shot learning settings~\cite{kojima2022large}. For each dataset, we randomly permute the order of nodes and edges in the serialized graphs and conduct 10 independent experiments to analyze performance variations across different orderings, leading to some counter-intuitive findings as shown in Figure~\ref{fig:boxplot}. Figure~\ref{fig:boxplot}(a) and Figure~\ref{fig:boxplot}(b) show that merely changing the order of nodes and edges in the questions causes performance fluctuations across all these models (both LLMs and Graph LLMs that have undergone embedding alignment and graph task fine-tuning). Additionally, as shown in Figure~\ref{fig:boxplot}(c) and Figure~\ref{fig:boxplot}(d), this phenomenon persists across models of the same architecture with different parameter scales, with no significant improvement as model parameters increase. Therefore, regarding Question I, it seems that current LLMs and Graph LLMs do not understand and process graphs well, as an ideal Graph LLM should maintain high performance regardless of the ordering used to represent the same graph. More detailed analysis and examples can be seen in Appendix~\ref{Appendix_exp_order}.

Furthermore, another question for Graph LLM is \textbf{Question II}: Does context based on randomly sampled partial neighborhoods limit model effectiveness? The current mainstream paradigm of LLM as predictor obtains subgraph representations by randomly sampling $n$ neighbors centered on target nodes and feeding them to LLMs~\cite{Ren2024ASO}, to accommodate LLMs' limited context length. Although studies on RAG make efforts on graph data compression~\cite{he2024g,hu2024grag}, they focus on retrieving external knowledge and enhancing LLM. Different from these studies, to our best knowledge, no work discusses how to determine optimal or relatively better input subgraphs from the provided graph rather than relying on random sampling. Random sampling may sample graph structures that are detrimental to graph learning, such as heterophily~\cite{Zhu2020BeyondHI,Pei2020GeomGCNGG}. This occurs because sampling nodes of different categories than the target node from its neighbors can lead to incorrect mixing of node features~\cite{Zhu2020BeyondHI}, making nodes indistinguishable and resulting in incorrect answers.

To address these issues, we propose a novel framework called GraphSOS (Graph \underline{S}ampling and \underline{O}rder \underline{S}election). For Question I, GraphSOS introduces an Order Selector Module that selects better sequence order for serialized graphs, which is then fed into the LLM along with the question. Order Selector Module ensures that LLMs receive relatively better-ordered inputs for any graph, thus maintaining performance. For Question II, we introduce a Subgraph Sampling Module before the graph enters the Order Selector Module, which samples subgraphs of target nodes from the graph. We train the random walk process of the Subgraph Sampling Module using concepts from reinforcement learning and preference learning~\cite{Brown2020LanguageMA,Rafailov2023DirectPO} to sample better subgraphs. Additionally, to ensure the model follows instructions and derives answers through analysis and logical reasoning of input graphs, we propose Graph Chain of Thought (Graph CoT) obtained through distillation. We use Graph CoT to enhance LLM's reasoning and zero-shot capabilities for graph tasks through instruction tuning.
Our contributions can be summarized as:

• We identify current Graph LLMs are sensitive to graph serialization order, and random subgraph sampling can mix node features from different categories incorrectly, affecting model performance.
    
• We propose GraphSOS, a novel framework that improves LLM graph processing via two key components: a Subgraph Sampling Module for optimal subgraph extraction and an Order Selector Module for better graph serialization order.
    
• We introduce Graph CoT obtained through distillation and employ instruction tuning to teach models to reason correctly about graph structures.
    
• We evaluate our model on node classification and graph question-answering tasks and analyze the impact of its components. We demonstrate GraphSOS’s superior performance in supervised and zero-shot graph learning settings.

\section{Preliminaries}
LLMs for graph data tasks can categorized into ``LLM as enhancer" and ``LLM as predictor" based on the role of LLM~\cite{Chen2023ExploringTP}. This paper focuses on LLM as predictor, which leverages LLM's reasoning capabilities to solve graph tasks, including graph Question-Answering (graph QA) and node classification on Text-Attributed Graph (TAG). A graph can formally represented as $\mathcal{G}(\mathcal{V},\mathcal{E},\mathbf{X})$, where $\mathcal{V}$ and $\mathcal{E}$ represent the sets of nodes and edges respectively. $\mathbf{X}$ denotes the feature matrix, where each row vector represents a node's attributes or feature information. For TAG node classification, each node corresponds to a text attribute in $\mathbf{X}$. For graph QA, $\mathcal{G}$ may not include meaningful $\mathbf{X}$, such as in shortest path problems.

LLM processes graph inputs in sequence form, thus requiring a graph encoding function to convert the graph into a sequence suitable for language models~\cite{Fatemi2023TalkLA}. The process of LLM obtaining answers can be represented as:
\begin{equation}\label{eq1}
    \mathcal{A}=f(g(\mathcal{G}),\mathcal{Q}),
\end{equation}
where $f(\cdot)$ formalizes the process of LLM obtaining answers from inputs, $\mathcal{Q}$ represents the user's question, $\mathcal{A}$ represents LLM's answer, and $g(\cdot)$ denotes the graph encoding function that converts graph $\mathcal{G}$ into a sequence, including serialization and possible subgraph sampling processes.

As shown in Table~\ref{tab:t1}, for TAG node classification tasks, we define $g(\mathcal{G})$ as the process of converting the node set $\mathcal{V}$ and node feature matrix $\mathbf{X}$ into a Feature List and converting the edge set $\mathcal{E}$ into an Edge List represented as pairs. For graph QA tasks, we convert the edge set $\mathcal{E}$ into pairs or triples based on specific task requirements. Since the graph QA tasks we study do not include node attributes, $g(\mathcal{G})$ for graph QA tasks does not contain a Feature List.

\begin{table}[tb]
\centering
\footnotesize
\caption{Construction of serialized text input for the graph.}
\label{tab:t1}
\begin{tabular}{@{}>{\centering\arraybackslash}p{2.5cm}|>{\centering\arraybackslash}p{5cm}@{}}
\toprule
Task & $g(\mathcal{G})$ \\ \midrule
TAG Node & Feature List: [Node 0 ... ], \\
Classification & Edge List: [(0, 1) (0, 2) ... ] \\ \midrule
Graph QA & Edge List: [(0, 1) (0, 2) ... ] \\ \midrule
Knowledge & Triple List: [(Lore, release\_year, \\
Graph QA & 2012) ... ] \\ \bottomrule
\end{tabular}
\end{table}

Our objectives are twofold. First, we optimize $g(\cdot)$ to enable LLM to obtain better-sampled subgraphs of target node $v$ in graph $\mathcal{G}$ and select relatively better serialization order representations for these subgraphs (in Sections~\ref{section:ssm} and~\ref{section:osm}). Secondly, we train and tune the LLM parameters to enhance the model's graph understanding and reasoning capabilities, to optimize $f(\cdot)$ (in Section~\ref{section:gcot}).

\section{Methodology}
\begin{figure}[t]
\centering
\includegraphics[width=0.49\textwidth]{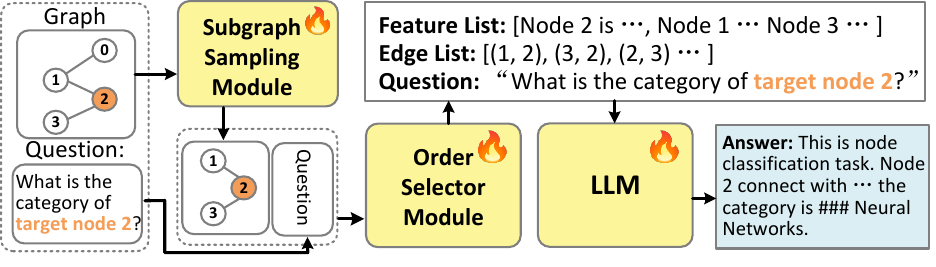} 
\caption{The overall framework of GraphSOS. Trainable components are highlighted in yellow and marked with flame icons. (a) Subgraph Sampling Module: samples and outputs a subgraph of a target node from graph $\mathcal{G}$. (b) Order Selector Module: takes the subgraph and user question, converts the subgraph into a text sequence and selects the sequence order. (c) LLM: generates answers based on question and serialized text representation of the graph.}
\label{pic: framework}
\end{figure}

In this section, we detail the proposed GraphSOS framework, with the overall architecture shown in Figure~\ref{pic: framework}. GraphSOS inputs both graph and question and generates natural language answers as output. The Subgraph Sampling Module takes graph $\mathcal{G}$ as input and aims to extract a subgraph of target node $v$ from $\mathcal{G}$. The Order Selector Module takes the subgraph and question as input, aiming to generate a natural language description of the subgraph and determine the sequence order of elements from both the Feature List and Edge List within the description. Finally, the instruction-tuned LLM generates answers in the Graph CoT answer format.
\begin{figure*}[t]
\centering
\includegraphics[width=0.95\textwidth]{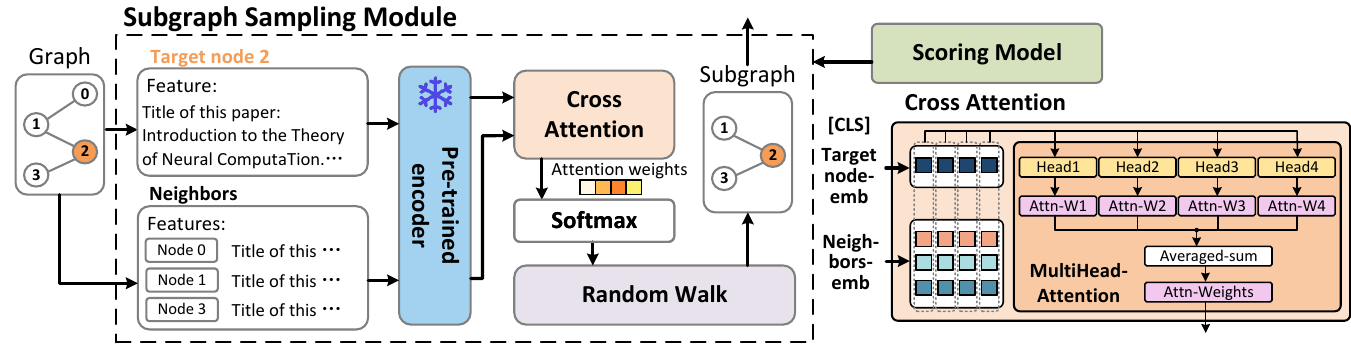} 
\caption{Internal details of the Subgraph Sampling Module (SSM). Frozen components are highlighted in blue and marked with snowflake icons.}
\label{pic: SSM}
\end{figure*}
\subsection{Subgraph Sampling Module}
\label{section:ssm}
In the Introduction, we address Question II: Does context based on randomly sampled partial neighborhoods limit model effectiveness? To improve this limitation, we design the Subgraph Sampling Module (SSM) to construct a subgraph $\mathcal{G}_{v}$ for target node $v$ from graph $\mathcal{G}$, rather than relying purely on random sampling. As shown in Figure~\ref{pic: SSM}, SSM obtains text attributes (i.e., features) of target node $v$ and its $k$-hop ($k=2$) neighbors, using a pre-trained encoder (PTE) to generate encoded features for each node. In our implementation, we use Bert~\cite{Devlin2019BERTPO} as the PTE, utilizing the [CLS] token embedding from Bert‘s last layer output as the representation for each node's features.

Next, we aim to use the [CLS] token embedding of target node $v$ as a query to compute its correlation with the [CLS] token embeddings of each neighbor node, guiding random walks to retain highly correlated neighbors in the sampled subgraph $\mathcal{G}_{v}$. We introduce scaled dot-product attention to compute correlation weights between $v$ and its neighbors, which is defined as~\cite{Vaswani2017AttentionIA}:
\vspace{-8pt}
\begin{equation}\label{eq2}
    Attention(Q,K,V)=\text{softmax}(\frac{QK^T}{\sqrt{d_k}})V,
\end{equation}
where $d_k$ is the embedding dimension, $Q$, $K$, and $V$ represent the embedding matrices for query, key, and value respectively. Here, $Q$ is the [CLS] token embedding vector of target node $v$, $K=V$ is the matrix composed of [CLS] token embeddings of $v$'s neighbors, and $\frac{QK^T}{\sqrt{d_k}}$ represents the correlation vector between query vector $Q$ and all vectors in key matrix $K$, defined as attention weights.

We then introduce multi-head attention based on dot-product attention. The [CLS] token embeddings of the target node $v$ and its neighbors are linearly mapped into $h$ ($h$=4) splits for multi-head cross-attention computation. In the (multi-head) cross-attention process, target node $v$'s embedding is mapped to queries, while neighbor nodes' embeddings are mapped to keys. Attention weights are computed based on query and key embeddings. Let $\mu$ be the split embeddings, the multi-head cross-attention process can be expressed as:
\begin{equation}\label{eq3}
    o_{attn} = \frac{1}{h}\sum_{i=1}^h head_i(\mu_i^v, (\mu_i^{u_1}, \mu_i^{u_2}, \dots, \mu_i^{u_n})),
\end{equation}
where $head_i$ represents attention weights from the $i$-th attention head, $h$ is the number of attention heads, $\mu_i^v$ is the $i$-th embedding component of target node $v$'s split embedding, $\mu_i^{u_n}$ is the $i$-th embedding component of neighbor node $u_n$'s split embedding, and $n$ is the total number of $v$'s $k$-hop neighbors. In our experiments, we use 4 attention heads ($h$=4) for multi-head cross-attention computation. The attention weights from multi-head attention guide random walks to retain highly correlated neighbors in sampled subgraph $\mathcal{G}_{v}$. Given maximum sample node count $n_\text{max}$, the random walk sampling process can be defined as:
\begin{equation}\label{eq4}
P(u_j|v) = o_{attn}[j],\
|\mathcal{V}_{\mathcal{G}_{v}}| \leq n_\text{max},
\end{equation}
where $o_{attn}[j]$ represents the $j$-th value in the attention weight vector, indicating the correlation between neighbor node $u_j$ and target node $v$, $P(u_j|v)$ is the probability of transitioning from node $v$ to neighbor node $u_j$ in the random walk process, $\mathcal{V}_{\mathcal{G}_{v}}$ represents the node set in the sampled subgraph, and $|\mathcal{V}_{\mathcal{G}_{v}}|$ represents its node count. The neighbor nodes sampled by random walk together with node $v$ form the subgraph $\mathcal{G}_{v}$.

\textbf{Scoring Model.} To train SSM, we draw inspiration from reinforcement learning and preference learning~\cite{Brown2020LanguageMA,Rafailov2023DirectPO}. We first construct 500 data instances from the text-attributed graph datasets Citeseer and Cornell (see Section~\ref{experimental_steup}), where each instance contains a target node and its 2-hop neighbors, and convert them into text descriptions using $g(\mathcal{G})$ as defined in Table~\ref{tab:t1}. Then, we construct positive and negative subgraph examples for each target node data point and train a Scoring Model using cross-entropy loss to score subgraphs. Based on task requirements, since we mainly focus on heterophily~\cite{Zhu2020BeyondHI,Pei2020GeomGCNGG}, positive examples are constructed as subgraphs with strong homophily, while negative examples are constructed as subgraphs with strong heterophily. We use Qwen 2.5-0.5B~\cite{qwen2_5} as the Scoring Model, requiring it to output 1 for positive examples and 0 for negative examples. To obtain continuous scores from the Scoring Model, we compute the softmax values of the probabilities that Qwen 2.5-0.5B predicts 0 or 1 for the first token, using the probability of predicting 1 from the softmax values as the model's score. This process can be described as:
\begin{equation}\label{eq5}
    P(y=1|g(\mathcal{G})) = \frac{e^{h_{\text{sc}}(g(\mathcal{G}))_{1,y=1}}}{e^{h_{\text{sc}}(g(\mathcal{G}))_{1,y=0}} + e^{h_{\text{sc}}(g(\mathcal{G}))_{1,y=1}}},
\end{equation}
where $h_{\text{sc}}$ represents the output logits from the Scoring Model's last layer, $y \in \{0,1\}$ indicates the preference label, $(\cdot)_{1,y=i}$ subscript represents taking the probability value of the first new token corresponding to digit $i$, and $P(y=1|g(\mathcal{G}))$ is the score given by the Scoring Model. We use this score to update SSM gradients, defining SSM's loss function as:
\begin{equation}\label{eq6}
    \mathcal{L_{SSM}} = \frac{1}{T}(1 - P(y=1|g(\mathcal{G})))^2,
\end{equation}
where $T$ is the temperature coefficient that adjusts the loss magnitude, and we set $T=5$ for smoother loss. It is worth noting that we use a scoring model to train the SSM, rather than directly using an LLM to replace SSM, because LLMs have limited input window sizes, while random walks in SSM have unlimited input windows. Moreover, constructing positive and negative example subgraphs for all target nodes for training would be computationally intensive. Through a Scoring Model trained on limited data, we can score any subgraph processed by SSM, greatly reducing training and data annotation costs.

\subsection{Order Selector Module}
\label{section:osm}
In the Introduction, we also address Question I: Do LLMs truly understand and process graph topological structures correctly? We demonstrate that both LLMs and Graph LLMs are sensitive to the order of elements in Feature List and Edge List, which deviates from ideal Graph LLMs, as a graph LLM should maintain high performance regardless of the sequence order in which its nodes and edges are described. To address this gap, we design the Order Selector Module (OSM). This module generates serialized representations of subgraphs produced by SSM and arranges the elements in Feature List and Edge List in optimized orders according to the user's questions. OSM ensures that for any input subgraph, its serialized representation is order-optimal or relatively optimal for model responses.

\begin{figure}[t]
\centering
\includegraphics[width=0.49\textwidth]{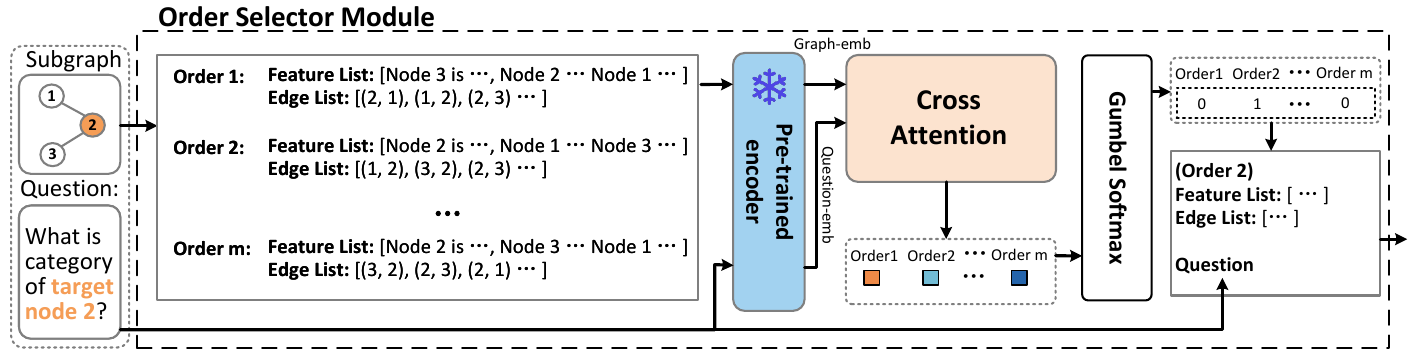} 
\caption{Internal details of the Order Selector Module (OSM). Frozen components are highlighted in blue and marked with snowflake icons.}
\label{pic: OSM}
\end{figure}

As shown in Figure~\ref{pic: OSM}, OSM receives a subgraph and serializes it into a natural language sequence consisting of Feature List and Edge List. Then, it generates $m$ order representations for this sequence, each being a random permutation of elements in the Feature List and Edge List. When $m$ can enumerate all possible permutations, OSM theoretically selects the optimal order, however, this would result in factorial computational complexity, which is unacceptable in terms of time cost. Therefore, in our experiments, we select a subset of all possible orders, setting $m=10$, allowing OSM to produce a fixed number of order permutations and obtain the relatively optimal order among them.

Next, the $m$ sequences and the user's question are input to PTE for encoding. We also use Bert~\cite{Devlin2019BERTPO} as the encoder, utilizing the [CLS] token embedding from its last layer as the representation for both the question and each ordered sequence. The cross-attention design is almost identical to SSM's, except that query $Q$ becomes the question's embedding vector, and key $K$ becomes the matrix composed of embeddings from $m$ ordered sequences. Cross-attention outputs attention weights, where each element represents the correlation weight between that ordered sequence and the question. We apply Gumbel Softmax~\cite{Jang2016CategoricalRW} to the attention weights to obtain a (one-hot) mask of length $m$, which selects a single optimal order from the $m$ ordered sequences as output. Gumbel Softmax is used to ensure model differentiability.

\textbf{Training OSM.} We train OSM and the LLM in Figure 2 as an end-to-end system, updating OSM parameters using the loss between the language model's output and target answers. Specifically, before training OSM, we first train the LLM through two-stage tuning (in Section~\ref{section:gcot}). Then, when training OSM, we freeze the LLM parameters and focus on updating OSM parameters. The loss function is constructed as:
\begin{equation}\label{eq7}
    \mathcal{L}_{OSM}(\pi_\theta) = -\sum_{i=1}^N \log \pi_\theta(y_i|x_i),
\end{equation}
where $\pi_\theta(y_i|x_i)$ is the conditional probability of LLM generating target output $y_i$ given input $x_i$, and $N$ is the number of training samples. During training, LLM parameters are frozen, and only OSM parameters are updated.

\subsection{Graph Chain-of-Thought Distillation}
\label{section:gcot}
Traditional graph learning, such as GNN, typically completes graph tasks in two steps~\cite{Li2015GatedGS,Kipf2016SemiSupervisedCW}: Step 1: Aggregate and update node features, Step 2: Make predictions based on aggregated features. However, although many studies input graphs along with questions to LLMs, LLMs often skip Step 1, ignoring graph structure and directly predicting answers, for example in node classification tasks~\cite{Tang2023GraphGPTGI} and graph question-answering tasks~\cite{Chen2024GraphWizAI}. This indicates that LLMs often complete graph tasks following incorrect graph reasoning steps.

\begin{figure}[t]
\centering
\includegraphics[width=0.49\textwidth]{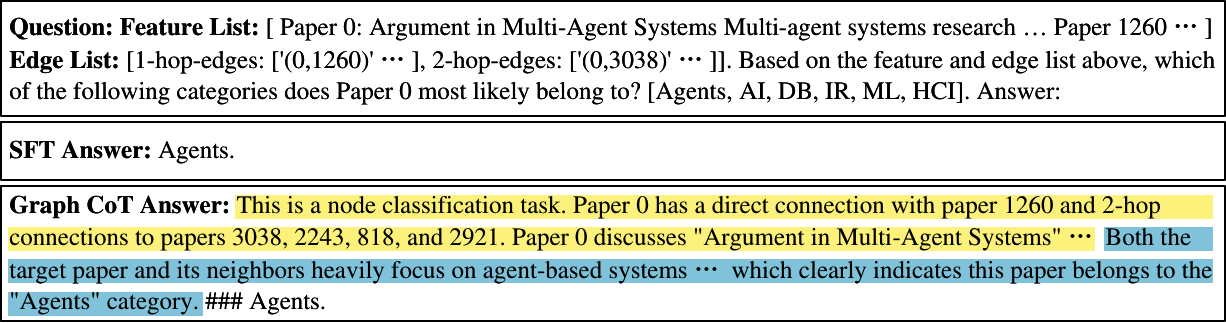} 
\caption{Constructing answers in Graph CoT format. Yellow highlights indicate analysis steps, while blue highlights show reasoning processes.}
\label{pic: GCoT}
\end{figure}

Inspired by Chain-of-Thought (CoT) research in language models~\cite{Wei2022ChainOT}, we define Graph CoT, decomposing graph task solution into two steps to maintain consistency with classical graph learning. Step 1: Analyze and understand features and structures on the graph, Step 2: Reason to obtain prediction results. Other CoT paradigms~\cite{Tang2023GraphGPTGI,Chen2024GraphWizAI} typically focus only on Step 2, contrasting with our Graph CoT. Recent research shows that closed-source models (like GPT-4) have strong capabilities in understanding and processing graph data~\cite{Guo2023GPT4GraphCL,Fatemi2023TalkLA}. We aim to distill the capabilities of large closed-source models into our small-parameter model through knowledge distillation. Following general steps for distilling knowledge from LLMs~\cite{Tang2023GraphGPTGI,Chen2024GraphWizAI}, we use GPT-4o to construct answers for the training dataset and construct prompts that make GPT-4o first analyze graph structure, then generate answers in CoT format, with examples shown in Figure~\ref{pic: GCoT}.

\textbf{Two-Stage Tuning for LLMs.} To enable language models to follow human instructions, instruction tuning is commonly used~\cite{Ouyang2022TrainingLM}. However, to further enable models to think and generate responses using Graph CoT, we define a two-stage tuning approach. Stage 1: Instruction tuning. Stage 2: Direct Preference Optimization (DPO)~\cite{Rafailov2023DirectPO}.
In the instruction tuning stage, the LLM is trained directly using the Question and SFT Answer format as shown in Figure~\ref{pic: GCoT}. We use LoRA~\cite{Hu2021LoRALA} to improve training efficiency. In the DPO phase, preference data is constructed using Graph CoT answers as winning responses and SFT answers as losing responses, encouraging the LLM to generate responses based on Graph CoT. Loss functions for the two stages are defined as:
\vspace{-4pt}
\begin{equation}\label{eq8}
    \mathcal{L}_{SFT}(\pi_\theta) = -\sum_{i=1}^N \log \pi_\theta(y_i|x_i),
\end{equation}
\begin{equation}\label{eq9}
\begin{aligned}
    \mathcal{L}_{DPO}(\pi_\theta) = -\mathbb{E}_{(x,y_w,y_l)\sim D} \log{\sigma[}&\beta\log(\frac{\pi_\theta(y_w|x)}{\pi_{ref}(y_w|x)}) \\
    - &\beta\log(\frac{\pi_\theta(y_l|x)}{\pi_{ref}(y_l|x)})]
\end{aligned}
\end{equation}
where $\pi_\theta$ is LLM parameters, $\pi_\theta(y_i|x_i)$ is the conditional probability of generating target output $y_i$ given input $x_i$, $\pi_{ref}$ represents the reference policy, which is the LLM parameter state after the first stage instruction tuning and before the second stage DPO training. $(x,y_w,y_l)$ is a triplet consisting of input question and its corresponding winning and losing answers, $\beta$ ($\beta=1$) is the temperature coefficient, $\sigma$ is the sigmoid function, and $D$ is the training dataset.

\section{Experiments}
In this section, we validate the proposed GraphSOS by addressing the following questions: \textbf{RQ1}: How does GraphSOS perform in supervised and zero-shot learning settings? \textbf{RQ2}: What are the contributions of each key component in GraphSOS to the overall performance? \textbf{RQ3}: How do the hyperparameters affect model performance? The detailed analysis of \textbf{RQ3} is provided in Appendix~\ref{Appendix_RQ3}.
\subsection{Experimental Setup}
\label{experimental_steup}
\textbf{Dataset.} We focus on Text-Attributed Graph (TAG) node classification and graph Question-Answering (graph QA) tasks. For TAG node classification, we use three homophily citation TAG datasets provided by Chen et al.~\cite{Chen2023ExploringTP}: Cora, Citeseer, and Pubmed, following their training and test set splits. Additionally, following Pei et al.~\cite{Pei2020GeomGCNGG}, we construct three high-heterophily text graph datasets: Texas, Wisconsin, and Cornell, from~\href{http://www.cs.cmu.edu/afs/cs.cmu.edu/project/theo-11/www/wwkb}{WebKB}. In these datasets, nodes represent webpages, and edges represent hyperlinks between them. Node features are webpage descriptions. Webpages are categorized into four classes: student, course, staff, and faculty, with training, validation, and test sets split in a 1:1:8 ratio.

For graph QA tasks, MetaQA~\cite{Zhang2017VariationalRF} is a movie knowledge graph QA dataset. We select 1,000 samples from the provided 2-hop test set (avoiding training set to prevent data contamination) and split them into training, validation, and test sets in a 1:1:8 ratio. For each question, we retrieve 1-hop and 2-hop triples centered around the target entity from the knowledge graph, with the number of triples per question ranging from 50 to 1,198. Furthermore, we use the graph reasoning QA dataset provided by Chen et al.~\cite{Chen2024GraphWizAI}, which includes 9 tasks with a total of 18.1k training samples and 400 test samples per task.

\textbf{Baseline Methods.} In our performance comparison, we consider various advanced methods for comprehensive evaluation: (i) The first category includes Graph Neural Networks (GNNs): We use two-layer GNNs, including GCN~\cite{Kipf2016SemiSupervisedCW}, GAT~\cite{velickovic2017graphan}, GraphSAGE~\cite{hamilton2017inductive}, NodeFormer~\cite{Wu2023NodeFormerAS}, etc. As well as GNNs with knowledge distillation: GKD~\cite{yang2022geometric}, GLNN~\cite{zhang2021graph}, and AdaGMLP~\cite{10.1145/3637528.3671699}. More baselines and results can be seen in Appendix~\ref{Appendix_Full_Results}. For TAG node classification tasks, node text attributes are encoded using BERT~\cite{Devlin2019BERTPO}, and the [CLS] token embedding is used as the node feature vector. For graph QA tasks, we sample one-dimensional features from a normal distribution as node features. (ii) The second category showcases state-of-the-art multi-task capable Graph LLMs, including GraphGPT-7B~\cite{Tang2023GraphGPTGI}, GraphWiz-LLaMa2-7B~\cite{Chen2024GraphWizAI}, and three variants of TALK-LIKE-A-GRAPH~\cite{Fatemi2023TalkLA} (GPT-Adjacency, GPT-Incident, and GPT-Expert), which utilize the closed-source GPT-3.5-turbo-16k~\cite{ouyang2022training} and thus do not participate in any training in our experiments. (iii) The third category consists of simple fine-tuned open-source LLMs, including Qwen 2.5-7B \cite{qwen2_5} and LLaMA 3-8B \cite{Dubey2024TheL3}. These models are fine-tuned using the SFT answer format shown in Figure \ref{pic: GCoT} for comparison with our approach.

\textbf{Implementation Details.}We construct GraphSOS using LLaMA 3-8B and Qwen 2.5-7B as backbone models. The experimental configuration details are in Appendix~\ref{Appendix_Experimental_Setup}.


\begin{table*}[t]
\centering
\small
\caption{Performance comparison (accuracy) on TAG node classification tasks under supervised and zero-shot settings.}
\label{tab: nodecls}
\resizebox{\textwidth}{!}{
\begin{tabular}{cccccccc}
\toprule
\multirow{3}{*}{}          & Dataset                & Citeseer           & Cora               & Pubmed             & Cornell            & Texas              & Wisconsin           \\
                           & Edge Hom.              & 0.78               & 0.81               & 0.80               & 0.26               & 0.25               & 0.33                \\
                           & Training Method        & SFT                & 0-shot             & 0-shot             & SFT                & 0-shot             & 0-shot              \\
\midrule
\multirow{7}{*}{\centering GNN} & GCN               & 70.7$\pm$0.4         & 13.9$\pm$3.2         & 26.3$\pm$2.8         & 47.4$\pm$3.9         & 8.9$\pm$7.7          & 21.2$\pm$21.4         \\
                           & GAT                    & 71.2$\pm$0.8         & 13.4$\pm$5.6         & 27.5$\pm$3.3         & 50.5$\pm$2.7         & 37.6$\pm$4.9         & 22.9$\pm$19.2         \\
                           & GraphSage & 70.9$\pm$0.6 & 24.2$\pm$14.1 & 25.8$\pm$3.0 & 48.9$\pm$3.2 & 29.5$\pm$6.8 & 23.5$\pm$20.1 \\
& NodeFormer & 70.5$\pm$0.8 & 14.5$\pm$4.0 & 26.1$\pm$3.1 & 75.5$\pm$1.4 & 30.1$\pm$6.6 & 22.8$\pm$20.3 \\
& GKD & 72.0$\pm$0.5 & 14.1$\pm$4.0 & 24.5$\pm$3.2 & 48.2$\pm$3.1 & 9.7$\pm$7.2 & 12.3$\pm$10.0 \\
& GLNN & 73.1$\pm$0.3 & 13.8$\pm$3.7 & 21.0$\pm$2.8 & 51.7$\pm$3.4 & 19.4$\pm$7.4 & 21.6$\pm$19.7 \\
& AdaGMLP & 72.8$\pm$0.4 & 14.1$\pm$3.5 & 11.5$\pm$7.9 & 71.2$\pm$1.3 & 20.1$\pm$7.2 & 22.0$\pm$19.5 \\
\midrule
\multirow{5}{*}{\centering Graph LLM} & GraphWiz    & 74.9$\pm$0.7         & 0.1$\pm$0.9          & 1.5$\pm$1.1          & 50.0$\pm$0.8         & 48.6$\pm$1.2         & 60.6$\pm$0.6         \\
                           & GraphGPT               & 53.2$\pm$1.3         & 9.1$\pm$0.5          & 70.1$\pm$1.4         & 49.8$\pm$0.7         & 52.3$\pm$0.9         & 60.0$\pm$1.1         \\
& GPT-Adjacency & 17.8$\pm$0.5 & 64.2$\pm$0.7 & 20.1$\pm$0.6 & 77.8$\pm$0.1 & 72.9$\pm$0.1 & 79.1$\pm$0.2 \\
& GPT-Incident & 18.6$\pm$0.4 & 65.4$\pm$0.3 & 20.2$\pm$0.7 & 78.2$\pm$0.0 & 73.1$\pm$0.1 & 80.2$\pm$0.0 \\
& GPT-Expert & 18.5$\pm$0.2 & 65.9$\pm$0.3 & 20.8$\pm$0.8 & 78.1$\pm$0.0 & 73.2$\pm$0.1 & 79.9$\pm$0.1 \\
\midrule
\multirow{4}{*}{\centering Qwen 2.5} & 1stage      & 38.4$\pm$0.8         & 36.8$\pm$1.2         & 20.2$\pm$0.6         & 71.9$\pm$1.5         & 64.4$\pm$0.4         & 79.1$\pm$0.9         \\
                           & \textbf{GraphSOS-2stage }        & 64.2$\pm$1.1         & 64.2$\pm$0.7         & 70.8$\pm$1.3         & 75.7$\pm$0.5         & 75.8$\pm$1.4         & 82.1$\pm$0.8         \\
                           & \textbf{GraphSOS-2stage-SSM}     & 65.3$\pm$0.6         & 65.4$\pm$1.4         & 72.3$\pm$0.9         & 77.3$\pm$1.2         & 76.9$\pm$0.3         & 83.5$\pm$1.0         \\
                           & \textbf{GraphSOS-2stage-SSM-OSM} & 69.7$\pm$0.8 & 66.3$\pm$0.6 & 73.9$\pm$1.2 & \textbf{80.1$\pm$0.6} & \textbf{78.6$\pm$0.9} & 84.9$\pm$0.5 \\
\midrule
\multirow{4}{*}{\centering LLaMA 3} & 1stage       & 74.5$\pm$1.2         & 9.7$\pm$0.8          & 7.6$\pm$0.5          & 76.5$\pm$1.3         & 68.5$\pm$0.7         & 79.5$\pm$1.4         \\
                           & \textbf{GraphSOS-2stage}         & 74.9$\pm$0.9         & 67.3$\pm$1.5         & 75.9$\pm$0.4         & 76.5$\pm$0.6         & 72.6$\pm$1.0         & 81.8$\pm$1.2         \\
                           & \textbf{GraphSOS-2stage-SSM}     & 75.3$\pm$0.3         & 68.5$\pm$1.1         & 76.1$\pm$1.4         & 78.9$\pm$0.9         & 74.3$\pm$0.5         & 83.0$\pm$0.8         \\
                           & \textbf{GraphSOS-2stage-SSM-OSM} & \textbf{77.0$\pm$0.5} & \textbf{70.5$\pm$0.8} & \textbf{77.6$\pm$0.7} & 79.5$\pm$0.9 & 76.5$\pm$0.7 & \textbf{85.2$\pm$0.6} \\
\bottomrule
\end{tabular}
}
\end{table*}

\begin{table*}[t]
\centering
\caption{Performance comparison (accuracy) on graph QA tasks under supervised and zero-shot settings.}
\label{tab: gqa}
\resizebox{\textwidth}{!}{
\begin{tabular}{cccccccccccc}
\toprule
                           & Dataset            & MetaQA    & cycle     & connect   & bipartite & topology  & shortest  & triangle  & flow      & hamilton   & subgraph   \\
                           & Training Method    & SFT       & SFT       & SFT       & SFT       & SFT       & SFT       & SFT       & SFT       & SFT/0-shot & SFT/0-shot \\
\midrule
\multirow{7}{*}{GNN}       & GCN                & -         & 82.5$\pm$1.3 & 73.0$\pm$0.8 & 81.3$\pm$1.1 & -         & 5.3$\pm$0.7  & 7.0$\pm$1.4  & 10.3$\pm$0.9 & -          & 62.0$\pm$1.2  \\
                           & GAT                & -         & 84.5$\pm$0.6 & 79.8$\pm$1.4 & 83.8$\pm$0.5 & -         & 7.3$\pm$1.2  & 7.3$\pm$0.8  & 11.8$\pm$1.5 & -          & 64.5$\pm$0.4  \\
                           & GraphSAGE & - & 83.8$\pm$1.1 & 78.5$\pm$1.2 & 82.9$\pm$0.8 & - & 7.0$\pm$0.9 & 7.5$\pm$1.1 & 11.2$\pm$1.2 & - & 63.8$\pm$0.8 \\
& NodeFormer & - & 83.5$\pm$0.9 & 79.0$\pm$1.0 & 82.7$\pm$1.0 & - & 6.8$\pm$1.0 & 8.7$\pm$0.9 & 13.0$\pm$1.3 & - & 58.2$\pm$0.7 \\
& GKD & - & 84.2$\pm$0.8 & 79.5$\pm$1.1 & 83.1$\pm$0.5 & - & 7.2$\pm$0.8 & 7.2$\pm$1.2 & 11.6$\pm$1.0 & - & 64.2$\pm$0.9 \\
& GLNN & - & 84.7$\pm$1.0 & 80.8$\pm$0.7 & \textbf{83.9$\pm$0.4} & - & 8.2$\pm$0.3 & 7.1$\pm$0.8 & 11.3$\pm$1.4 & - & 63.5$\pm$1.1 \\
& AdaGMLP & - & 84.5$\pm$0.9 & 80.5$\pm$0.8 & 83.6$\pm$0.5 & - & 8.0$\pm$0.4 & 7.3$\pm$0.9 & 11.5$\pm$1.3 & - & 63.8$\pm$1.0 \\
\midrule
\multirow{5}{*}{Graph LLM} & GraphWiz           & 35.3$\pm$1.9  & 70.0$\pm$1.1 & 89.8$\pm$0.3 & 73.3$\pm$1.4 & 16.3$\pm$0.7 & 12.8$\pm$1.0 & 24.0$\pm$0.6 & 28.3$\pm$1.3 & 39.0$\pm$0.8  & 70.3$\pm$1.5  \\
                           & GraphGPT           & 32.9$\pm$1.2 & 72.8$\pm$0.4 & 83.5$\pm$1.5 & 66.8$\pm$0.7 & 0.0$\pm$0.0  & 9.3$\pm$0.5  & 23.3$\pm$1.3 & 13.3$\pm$0.8 & 31.8$\pm$1.4  & 59.8$\pm$0.6  \\
& GPT-Adjacency & 86.9$\pm$1.1 & 81.2$\pm$0.9 & 89.5$\pm$1.0 & 77.8$\pm$0.8 & 70.1$\pm$0.7 & 24.2$\pm$0.7 & 34.8$\pm$1.2 & 36.2$\pm$0.9 & 41.5$\pm$1.1 & 65.1$\pm$0.7 \\
& GPT-Incident & 86.8$\pm$1.1 & 84.9$\pm$0.8 & 90.3$\pm$0.9 & 79.1$\pm$0.7 & \textbf{72.3$\pm$0.8} & 14.5$\pm$0.9 & 25.7$\pm$0.8 & 37.1$\pm$1.1 & 41.2$\pm$0.9 & 66.8$\pm$1.0 \\
& GPT-Expert & 86.9$\pm$1.9 & 81.8$\pm$1.0 & 89.8$\pm$1.1 & 78.2$\pm$0.9 & 70.0$\pm$0.7 & 24.9$\pm$0.6 & 35.1$\pm$1.0 & 36.5$\pm$0.8 & 41.1$\pm$1.2 & 66.8$\pm$0.8 \\
\midrule
\multirow{3}{*}{Qwen 2.5}  & 1stage             & 40.8$\pm$3.5 & 79.3$\pm$1.4 & 88.0$\pm$0.7 & 73.3$\pm$1.2 & 0.0$\pm$0.0  & 10.3$\pm$1.5 & 24.0$\pm$0.9 & 27.3$\pm$0.4 & 31.8$\pm$1.1  & 59.8$\pm$0.8  \\
                           & \textbf{GraphSOS-2stage}     & 80.3$\pm$6.8 & 79.5$\pm$0.6 & 88.0$\pm$1.1 & 74.8$\pm$0.9 & 15.8$\pm$1.4 & 16.8$\pm$0.7 & 21.8$\pm$1.2 & 20.0$\pm$1.5 & 32.8$\pm$0.5  & 68.8$\pm$1.0  \\
                           & \textbf{GraphSOS-2stage-OSM} & 86.9$\pm$3.5 & 80.4$\pm$1.2 & 89.3$\pm$0.6 & 77.1$\pm$0.9 & 16.7$\pm$0.8 & 16.9$\pm$1.1 & 24.2$\pm$0.9 & 26.8$\pm$1.2 & 36.3$\pm$1.2  & 68.9$\pm$0.5  \\
\midrule
\multirow{3}{*}{LLaMA 3}   & 1stage             & 46.4$\pm$3.4 & 83.5$\pm$0.7 & 90.0$\pm$1.2 & 78.5$\pm$0.5 & 0.0$\pm$0.0  & 14.8$\pm$0.8 & 35.3$\pm$1.1 & 25.8$\pm$0.6 & 31.8$\pm$1.3  & 62.5$\pm$0.9  \\
                           & \textbf{GraphSOS-2stage }    & 83.3$\pm$1.1 & 89.8$\pm$1.3 & 92.8$\pm$0.8 & 79.3$\pm$1.4 & 17.3$\pm$0.4 & 24.5$\pm$1.1 & 37.5$\pm$0.5 & 38.5$\pm$1.2 & 41.0$\pm$0.7  & 69.5$\pm$1.5  \\
                           & \textbf{GraphSOS-2stage-OSM} & \textbf{89.6$\pm$1.0} & \textbf{92.7$\pm$0.7} & \textbf{93.4$\pm$1.2} & 80.2$\pm$0.9 & 18.8$\pm$0.8 & \textbf{26.3$\pm$0.5} & \textbf{41.2$\pm$1.3} & \textbf{40.7$\pm$1.0} & \textbf{42.7$\pm$0.9}  & \textbf{72.9$\pm$0.8} \\
\bottomrule
\end{tabular}
}
\end{table*}

\subsection{Overall Performance of GraphSOS (RQ1)}
We evaluate the models on TAG node classification and graph QA tasks to assess their performance in both supervised and zero-shot settings. The overall performance is shown in Table~\ref{tab: nodecls} and Table~\ref{tab: gqa}. Supervised tasks are labeled as SFT, indicating that models are trained on a dataset's training split and evaluated on its corresponding test split (e.g., training on Citeseer's training data and evaluating on Citeseer's test set). Zero-shot learning is labeled as 0-shot, indicating that models trained on other datasets are directly tested on the target dataset without additional training (e.g., training on Citeseer and testing on Cora dataset). ``1stage" represents a simple baseline where open-source LLMs are fine-tuned using SFT answers shown in Figure~\ref{pic: GCoT}. ``2stage" indicates models trained using the two-stage tuning approach described in Section~\ref{section:gcot}. ``SSM" indicates the use of Subgraph Sampling Module instead of random selection for sampling target node neighbors from the graph, and ``OSM" indicates the use of Order Selector Module for selecting element order in serialized graphs. As complete graph structures are necessary for graph QA tasks, SSM or random sampling is not used for graph QA tasks; instead, complete graphs serve as input.

\textbf{TAG Node Classification.} In Table~\ref{tab: nodecls}, we introduce edge homophily measure~\cite{AbuElHaija2019MixHopHG}, formally defined as: $H_{edge}(\mathcal{G}) = \frac{|\{e_{uv} | e_{uv} \in \mathcal{E}, Z_{u,:} = Z_{v,:}\}|}{\mathcal{|E|}}$, representing the proportion of edges connecting nodes of the same class, where $Z_{u,:} = Z_{v,:}$ indicates nodes $u$ and $v$ belong to the same category. Each dataset's $H_{edge}(\mathcal{G})$ is labeled in the Edge Hom. row, with values closer to 0 indicating stronger heterophily in $\mathcal{G}$. All models are supervised trained only on Citeseer and Cornell training sets, with zero-shot learning on the remaining datasets. Results show that in SFT settings, LLaMA 3-GraphSOS-2stage-SSM-OSM outperforms GNNs on Citeseer and Cornell, particularly on heterophily graph Cornell, demonstrating strong heterophily graph learning capabilities. In the zero-shot learning setting, all variants of GraphSOS demonstrate consistent performance advantages, achieving 1-5 times higher accuracy compared to other baselines.

\textbf{Graph QA.} In Table~\ref{tab: gqa}, GNNs are supervised trained except for MetaQA, topology, and hamilton, as GNNs do not support path reasoning. Datasets labeled as SFT indicate supervised fine-tuning using their respective training sets. Specifically, as the most challenging tasks in graph reasoning QA~\cite{Chen2024GraphWizAI}, hamilton and subgraph use zero-shot settings for all LLM-based methods except GraphWiz to test zero-shot learning capabilities, while GraphWiz, designed and trained specifically for graph reasoning QA, is fine-tuned on training sets of all graph QA datasets for comparison. Experimental results show that in SFT settings, GraphSOS leads performance on almost all datasets compared to all baselines while maintaining stable performance. Additionally, ``2stage" models typically perform better than ``1stage" indicating Graph CoT benefits graph reasoning tasks. In zero-shot learning settings (hamilton and subgraph), despite no supervised training, GraphSOS achieves performance close to or even surpassing supervised-trained GraphWiz, demonstrating GraphSOS's ability to understand and reason about graphs learned from other tasks.

\begin{table}[t]
\centering
\caption{Ablation study on SSM, OSM, and Graph CoT of GraphSOS.}
\label{tab: ablation}
\resizebox{0.49\textwidth}{!}{
\begin{tabular}{ccccccc}
\toprule
              & Citeseer  & Cora      & Texas     & MetaQA    & cycle     & subgraph  \\
\midrule
w/o SSM       & 74.9$\pm$1.2 & 67.5$\pm$0.8 & 69.6$\pm$1.4 & -         & -         & -         \\
w/o OSM       & 75.3$\pm$0.3 & 68.5$\pm$1.1 & 74.3$\pm$0.5 & 83.3$\pm$1.1 & 89.8$\pm$1.3 & 69.5$\pm$1.5 \\
w/o Graph CoT & 71.6$\pm$1.4 & 9.3$\pm$0.6  & 69.3$\pm$1.2 & 47.7$\pm$0.8 & 85.2$\pm$1.5 & 64.9$\pm$0.4 \\
\textbf{GraphSOS}       & \textbf{77.0$\pm$0.5} & \textbf{70.5$\pm$0.8} & \textbf{76.5$\pm$0.7} & \textbf{89.6$\pm$1.0} & \textbf{92.7$\pm$0.7} & \textbf{72.9$\pm$0.8} \\
\bottomrule
\end{tabular}
}
\end{table}

\subsection{Module Ablation Study (RQ2)}
We conduct ablation studies to explore the individual contributions of different components in our proposed GraphSOS. Using LLaMA 3-8B as the base model, results are shown in Table~\ref{tab: ablation}.

\textbf{Effect of Subgraph Sampling.} We study the benefits of Subgraph Sampling Module using the ``w/o SSM" variant. In this variant, we directly randomly sample neighbors of target nodes for node classification across three datasets, without using SSM for neighbor sampling. Results in Table~\ref{tab: ablation} show that GraphSOS with SSM outperforms its variant using random sampling. This indicates SSM's ability to select task-relevant subgraphs, especially for heterophily graphs (e.g., Texas), where SSM can reduce the selection of dissimilar neighbors to target nodes. As shown in Figure~\ref{pic: ablation_ssm} in Appendix~\ref{Appendix_Ablation_ Results}, compared to the ``w/o SSM" variant, SSM samples a notably higher proportion of same-class neighbors. Research in graph learning suggests this benefits node classification tasks~\cite{McPherson2001BirdsOA,Battaglia2018RelationalIB}.

\textbf{Effect of Order Selection.} We study the benefits of Order Selector Module in selecting element order in Feature List and Edge List of serialized graphs using the ``w/o OSM" variant. In this variant, we randomly arrange elements in Feature List and Edge List and conduct experiments on node classification and graph QA tasks. The results in Table~\ref{tab: ablation} demonstrate that GraphSOS with OSM outperforms its variants, indicating that the context ordering selected by OSM helps LLMs understand and reason about graphs, thereby ensuring high performance. Additionally, as shown in Figure~\ref{pic: ablation_osm} in Appendix~\ref{Appendix_Ablation_ Results}, in statistical results of 10 independent experiments with randomly permuted element orders in Feature List and Edge List for each dataset, GraphSOS with OSM shows smaller performance fluctuation, indicating OSM's ability to suppress LLM's order sensitivity.

\textbf{Effect of Graph CoT.} We study the benefits of using DPO to teach models to generate Graph CoT answers during two-stage tuning using the ``w/o Graph CoT" variant. In this variant, we use instruction tuning with SFT answers from Figure~\ref{pic: GCoT} and conduct experiments on node classification and graph QA tasks. Results in Table~\ref{tab: ablation} show that GraphSOS with two-stage tuning demonstrates consistent performance advantages over single-stage instruction tuning models. Notably, this performance advantage is also evident in zero-shot learning settings on the subgraph dataset, indicating Graph CoT helps models transfer reasoning abilities learned from other tasks to specific tasks, demonstrating zero-shot reasoning and generalization capabilities.

\section{Conclusion}
This paper introduces GraphSOS to address design deficiencies in Graph LLMs. We focus on two major issues: order sensitivity to serialized graphs and random subgraph sampling as input. We propose the Order Selector Module and Subgraph Sampling Module as solutions. We also introduce Graph CoT to enhance LLMs' graph reasoning. GraphSOS shows improved performance on graph tasks in both supervised and zero-shot learning settings.


\section{Impact Statement}

This work advances graph learning capabilities of Large Language Models through improved sampling, ordering, and reasoning mechanisms. While our proposed Graph Chain of Thought (Graph CoT) approach enhances model interpretability and performance by making reasoning steps explicit, we acknowledge several important considerations regarding its societal and ethical implications.

The primary ethical consideration stems from the inherent limitations of base LLMs that GraphSOS builds upon. Despite the structured reasoning paths, LLMs may still exhibit hallucinations or generate incorrect logical steps, which could propagate through the graph analysis process. This is particularly concerning in high-stakes applications like social network analysis or recommendation systems, where faulty reasoning could lead to biased or harmful decisions.

To mitigate these risks, we recommend:
\begin{enumerate}
    \item Using more robust base models with demonstrated reliability.
    \item Creating diverse and carefully curated Graph CoT training data that emphasizes accurate reasoning.
    \item Implementing validation mechanisms to verify the logical consistency of generated reasoning chains.
    \item Maintaining human oversight in critical applications.
\end{enumerate}

Additionally, while Graph CoT improves transparency, it should not be considered a complete solution for model interpretability. The reasoning chains, while human-readable, may still reflect underlying biases present in the training data or model architecture.

We believe the benefits of enhanced graph understanding and explicit reasoning outweigh these risks when proper precautions are taken. Our work makes LLMs for graph tasks more reliable and interpretable, crucial for their responsible deployment in real-world applications.


\bibliography{example_paper}

\begin{thebibliography}{52}
\providecommand{\natexlab}[1]{#1}
\providecommand{\url}[1]{\texttt{#1}}
\expandafter\ifx\csname urlstyle\endcsname\relax
  \providecommand{\doi}[1]{doi: #1}\else
  \providecommand{\doi}{doi: \begingroup \urlstyle{rm}\Url}\fi

\bibitem[Abu-El-Haija et~al.(2019)Abu-El-Haija, Perozzi, Kapoor, and et~al.]{AbuElHaija2019MixHopHG}
Abu-El-Haija, S., Perozzi, B., Kapoor, A., and et~al.
\newblock Mixhop: Higher-order graph convolutional architectures via sparsified neighborhood mixing.
\newblock In \emph{International Conference on Machine Learning}, 2019.

\bibitem[An et~al.(2024)An, Ma, Lin, Zheng, and Lou]{An2024MakeYL}
An, S., Ma, Z., Lin, Z., Zheng, N., and Lou, J.-G.
\newblock Make your llm fully utilize the context.
\newblock \emph{ArXiv}, abs/2404.16811, 2024.

\bibitem[Bai et~al.(2023)Bai, Bai, Chu, and et~al.]{Bai2023QwenTR}
Bai, J., Bai, S., Chu, Y., and et~al.
\newblock Qwen technical report.
\newblock \emph{ArXiv}, abs/2309.16609, 2023.

\bibitem[Battaglia et~al.(2018)Battaglia, Hamrick, Bapst, Sanchez-Gonzalez, Zambaldi, Malinowski, Tacchetti, Raposo, Santoro, Faulkner, Çaglar G{\"u}lçehre, Song, Ballard, Gilmer, Dahl, Vaswani, Allen, Nash, Langston, Dyer, Heess, Wierstra, Kohli, Botvinick, Vinyals, Li, and Pascanu]{Battaglia2018RelationalIB}
Battaglia, P.~W., Hamrick, J.~B., Bapst, V., Sanchez-Gonzalez, A., Zambaldi, V.~F., Malinowski, M., Tacchetti, A., Raposo, D., Santoro, A., Faulkner, R., Çaglar G{\"u}lçehre, Song, H.~F., Ballard, A.~J., Gilmer, J., Dahl, G.~E., Vaswani, A., Allen, K.~R., Nash, C., Langston, V., Dyer, C., Heess, N. M.~O., Wierstra, D., Kohli, P., Botvinick, M.~M., Vinyals, O., Li, Y., and Pascanu, R.
\newblock Relational inductive biases, deep learning, and graph networks.
\newblock \emph{ArXiv}, abs/1806.01261, 2018.
\newblock URL \url{https://api.semanticscholar.org/CorpusID:46935302}.

\bibitem[Brown et~al.(2020)Brown, Mann, Ryder, and et~al.]{Brown2020LanguageMA}
Brown, T.~B., Mann, B., Ryder, N., and et~al.
\newblock Language models are few-shot learners.
\newblock \emph{ArXiv}, abs/2005.14165, 2020.

\bibitem[Chen et~al.(2024)Chen, Li, Tang, and Li]{Chen2024GraphWizAI}
Chen, N., Li, Y., Tang, J., and Li, J.
\newblock Graphwiz: An instruction-following language model for graph computational problems.
\newblock In \emph{Knowledge Discovery and Data Mining}, 2024.

\bibitem[Chen et~al.(2023)Chen, Mao, Li, Jin, Wen, Wei, Wang, Yin, Fan, Liu, and Tang]{Chen2023ExploringTP}
Chen, Z., Mao, H., Li, H., Jin, W., Wen, H., Wei, X., Wang, S., Yin, D., Fan, W., Liu, H., and Tang, J.
\newblock Exploring the potential of large language models (llms)in learning on graphs.
\newblock \emph{ACM SIGKDD Explorations Newsletter}, 25:\penalty0 42 -- 61, 2023.

\bibitem[Dai et~al.(2022)Dai, Zhou, Guo, and Wang]{pmlr-v198-dai22b}
Dai, E., Zhou, S., Guo, Z., and Wang, S.
\newblock Label-wise graph convolutional network for heterophilic graphs.
\newblock In Rieck, B. and Pascanu, R. (eds.), \emph{Proceedings of the First Learning on Graphs Conference}, volume 198 of \emph{Proceedings of Machine Learning Research}, pp.\  26:1--26:21. PMLR, 09--12 Dec 2022.
\newblock URL \url{https://proceedings.mlr.press/v198/dai22b.html}.

\bibitem[Defferrard et~al.(2016)Defferrard, Bresson, and Vandergheynst]{defferrard2016convolutional}
Defferrard, M., Bresson, X., and Vandergheynst, P.
\newblock Convolutional neural networks on graphs with fast localized spectral filtering.
\newblock \emph{Advances in neural information processing systems}, 29, 2016.

\bibitem[Devlin et~al.(2019)Devlin, Chang, Lee, and Toutanova]{Devlin2019BERTPO}
Devlin, J., Chang, M.-W., Lee, K., and Toutanova, K.
\newblock Bert: Pre-training of deep bidirectional transformers for language understanding.
\newblock In \emph{North American Chapter of the Association for Computational Linguistics}, 2019.

\bibitem[Dubey et~al.(2024)Dubey, Jauhri, Pandey, and et~al.]{Dubey2024TheL3}
Dubey, A., Jauhri, A., Pandey, A., and et~al.
\newblock The llama 3 herd of models.
\newblock \emph{ArXiv}, abs/2407.21783, 2024.

\bibitem[Fan et~al.(2019)Fan, Ma, Li, and et~al.]{fan2019graph}
Fan, W., Ma, Y., Li, Q., and et~al.
\newblock Graph neural networks for social recommendation.
\newblock In \emph{The world wide web conference}, pp.\  417--426, 2019.

\bibitem[Fang et~al.(2024)Fang, Guo, Zhou, Ma, Zhang, and Feng]{Fang2024LLaMAOmniSS}
Fang, Q., Guo, S., Zhou, Y., Ma, Z., Zhang, S., and Feng, Y.
\newblock Llama-omni: Seamless speech interaction with large language models.
\newblock \emph{ArXiv}, abs/2409.06666, 2024.

\bibitem[Fatemi et~al.(2023)Fatemi, Halcrow, and Perozzi]{Fatemi2023TalkLA}
Fatemi, B., Halcrow, J.~J., and Perozzi, B.
\newblock Talk like a graph: Encoding graphs for large language models.
\newblock \emph{ArXiv}, abs/2310.04560, 2023.

\bibitem[Guo et~al.(2023)Guo, Du, and Liu]{Guo2023GPT4GraphCL}
Guo, J., Du, L., and Liu, H.
\newblock Gpt4graph: Can large language models understand graph structured data ? an empirical evaluation and benchmarking.
\newblock \emph{ArXiv}, abs/2305.15066, 2023.

\bibitem[Hamilton et~al.(2017)Hamilton, Ying, and Leskovec]{hamilton2017inductive}
Hamilton, W., Ying, Z., and Leskovec, J.
\newblock Inductive representation learning on large graphs.
\newblock \emph{Advances in neural information processing systems}, 30, 2017.

\bibitem[He et~al.(2024)He, Tian, Sun, Chawla, Laurent, LeCun, Bresson, and Hooi]{he2024g}
He, X., Tian, Y., Sun, Y., Chawla, N.~V., Laurent, T., LeCun, Y., Bresson, X., and Hooi, B.
\newblock G-retriever: Retrieval-augmented generation for textual graph understanding and question answering.
\newblock \emph{arXiv preprint arXiv:2402.07630}, 2024.

\bibitem[Hu et~al.(2022)Hu, Shen, Wallis, and et~al.]{Hu2021LoRALA}
Hu, E.~J., Shen, Y., Wallis, P., and et~al.
\newblock Lo{RA}: Low-rank adaptation of large language models.
\newblock In \emph{International Conference on Learning Representations}, 2022.

\bibitem[Hu et~al.(2024)Hu, Lei, Zhang, Pan, Ling, and Zhao]{hu2024grag}
Hu, Y., Lei, Z., Zhang, Z., Pan, B., Ling, C., and Zhao, L.
\newblock Grag: Graph retrieval-augmented generation.
\newblock \emph{arXiv preprint arXiv:2405.16506}, 2024.

\bibitem[Jang et~al.(2016)Jang, Gu, and Poole]{Jang2016CategoricalRW}
Jang, E., Gu, S.~S., and Poole, B.
\newblock Categorical reparameterization with gumbel-softmax.
\newblock \emph{ArXiv}, abs/1611.01144, 2016.

\bibitem[Jiang et~al.(2019)Jiang, Xu, Araki, and Neubig]{Jiang2019HowCW}
Jiang, Z., Xu, F.~F., Araki, J., and Neubig, G.
\newblock How can we know what language models know?
\newblock \emph{Transactions of the Association for Computational Linguistics}, 8:\penalty0 423--438, 2019.

\bibitem[Kipf \& Welling(2016)Kipf and Welling]{Kipf2016SemiSupervisedCW}
Kipf, T. and Welling, M.
\newblock Semi-supervised classification with graph convolutional networks.
\newblock \emph{ArXiv}, abs/1609.02907, 2016.

\bibitem[Kojima et~al.(2022)Kojima, Gu, Reid, and et~al.]{kojima2022large}
Kojima, T., Gu, S.~S., Reid, M., and et~al.
\newblock Large language models are zero-shot reasoners.
\newblock \emph{Advances in neural information processing systems}, 35:\penalty0 22199--22213, 2022.

\bibitem[Kumar et~al.(2022)Kumar, Mallik, and Khetarpal]{kumar2022influence}
Kumar, S., Mallik, A., and Khetarpal, A.
\newblock Influence maximization in social networks using graph embedding and graph neural network.
\newblock \emph{Information Sciences}, 607:\penalty0 1617--1636, 2022.

\bibitem[Li et~al.(2023)Li, Li, Savarese, and Hoi]{Li2023BLIP2BL}
Li, J., Li, D., Savarese, S., and Hoi, S. C.~H.
\newblock Blip-2: Bootstrapping language-image pre-training with frozen image encoders and large language models.
\newblock In \emph{International Conference on Machine Learning}, 2023.

\bibitem[Li et~al.(2015)Li, Tarlow, Brockschmidt, and Zemel]{Li2015GatedGS}
Li, Y., Tarlow, D., Brockschmidt, M., and Zemel, R.
\newblock Gated graph sequence neural networks.
\newblock \emph{arXiv preprint arXiv:1511.05493}, 2015.

\bibitem[Liu et~al.(2024)Liu, Cai, Xu, and et~al.]{liu2024rumor}
Liu, T., Cai, Q., Xu, C., and et~al.
\newblock Rumor detection with a novel graph neural network approach.
\newblock \emph{Academic Journal of Science and Technology}, 10\penalty0 (1):\penalty0 305--310, 2024.

\bibitem[Lu et~al.(2024)Lu, Guan, Zhao, and Yang]{10.1145/3637528.3671699}
Lu, W., Guan, Z., Zhao, W., and Yang, Y.
\newblock Adagmlp: Adaboosting gnn-to-mlp knowledge distillation.
\newblock In \emph{Proceedings of the 30th ACM SIGKDD Conference on Knowledge Discovery and Data Mining}, KDD '24, pp.\  2060–2071, New York, NY, USA, 2024. Association for Computing Machinery.
\newblock ISBN 9798400704901.
\newblock \doi{10.1145/3637528.3671699}.
\newblock URL \url{https://doi.org/10.1145/3637528.3671699}.

\bibitem[Lu et~al.(2022)Lu, Bartolo, Moore, Riedel, and Stenetorp]{Lu2021FantasticallyOP}
Lu, Y., Bartolo, M., Moore, A., Riedel, S., and Stenetorp, P.
\newblock Fantastically ordered prompts and where to find them: Overcoming few-shot prompt order sensitivity.
\newblock In \emph{Proceedings of the 60th Annual Meeting of the Association for Computational Linguistics (Volume 1: Long Papers)}, pp.\  8086--8098, 2022.

\bibitem[Luo et~al.(2024)Luo, Liu, and Pan]{luo2024collaborative}
Luo, T., Liu, Y., and Pan, S.~J.
\newblock Collaborative sequential recommendations via multi-view gnn-transformers.
\newblock \emph{ACM Transactions on Information Systems}, 2024.

\bibitem[McPherson et~al.(2001)McPherson, Smith-Lovin, and Cook]{McPherson2001BirdsOA}
McPherson, M., Smith-Lovin, L., and Cook, J.~M.
\newblock Birds of a feather: Homophily in social networks.
\newblock \emph{Review of Sociology}, 27:\penalty0 415--444, 2001.

\bibitem[Ouyang et~al.(2022{\natexlab{a}})Ouyang, Wu, Jiang, Almeida, Wainwright, Mishkin, Zhang, Agarwal, Slama, Ray, et~al.]{ouyang2022training}
Ouyang, L., Wu, J., Jiang, X., Almeida, D., Wainwright, C., Mishkin, P., Zhang, C., Agarwal, S., Slama, K., Ray, A., et~al.
\newblock Training language models to follow instructions with human feedback.
\newblock \emph{Advances in neural information processing systems}, 35:\penalty0 27730--27744, 2022{\natexlab{a}}.

\bibitem[Ouyang et~al.(2022{\natexlab{b}})Ouyang, Wu, Jiang, and et~al.]{Ouyang2022TrainingLM}
Ouyang, L., Wu, J., Jiang, X., and et~al.
\newblock Training language models to follow instructions with human feedback.
\newblock \emph{ArXiv}, abs/2203.02155, 2022{\natexlab{b}}.

\bibitem[Pareja et~al.(2020)Pareja, Domeniconi, Chen, and et~al.]{pareja2020evolvegcn}
Pareja, A., Domeniconi, G., Chen, J., and et~al.
\newblock Evolvegcn: Evolving graph convolutional networks for dynamic graphs.
\newblock In \emph{Proceedings of the AAAI conference on artificial intelligence}, volume~34, pp.\  5363--5370, 2020.

\bibitem[Pei et~al.(2020)Pei, Wei, Chang, Lei, and Yang]{Pei2020GeomGCNGG}
Pei, H., Wei, B., Chang, K. C.-C., Lei, Y., and Yang, B.
\newblock Geom-gcn: Geometric graph convolutional networks.
\newblock \emph{ArXiv}, abs/2002.05287, 2020.

\bibitem[Rafailov et~al.(2024)Rafailov, Sharma, Mitchell, and et~al.]{Rafailov2023DirectPO}
Rafailov, R., Sharma, A., Mitchell, E., and et~al.
\newblock Direct preference optimization: Your language model is secretly a reward model.
\newblock \emph{Advances in Neural Information Processing Systems}, 36, 2024.

\bibitem[Ren et~al.(2024)Ren, Tang, Yin, Chawla, and Huang]{Ren2024ASO}
Ren, X., Tang, J., Yin, D., Chawla, N.~V., and Huang, C.
\newblock A survey of large language models for graphs.
\newblock In \emph{Knowledge Discovery and Data Mining}, 2024.

\bibitem[Tan et~al.(2024)Tan, Chu, Li, and Mo]{tan2024order}
Tan, Z., Chu, X., Li, W., and Mo, T.
\newblock Order matters: Exploring order sensitivity in multimodal large language models.
\newblock \emph{arXiv preprint arXiv:2410.16983}, 2024.

\bibitem[Tang et~al.(2024)Tang, Yang, Wei, and et~al.]{Tang2023GraphGPTGI}
Tang, J., Yang, Y., Wei, W., and et~al.
\newblock Graphgpt: Graph instruction tuning for large language models.
\newblock In \emph{Proceedings of the 47th International ACM SIGIR Conference on Research and Development in Information Retrieval}, pp.\  491--500, 2024.

\bibitem[Team(2024)]{qwen2_5}
Team, Q.
\newblock Qwen2.5: A party of foundation models, September 2024.
\newblock URL \url{https://qwenlm.github.io/blog/qwen2.5/}.

\bibitem[Touvron et~al.(2023)Touvron, Lavril, Izacard, Martinet, Lachaux, Lacroix, Rozi{\`e}re, Goyal, Hambro, Azhar, Rodriguez, Joulin, Grave, and Lample]{Touvron2023LLaMAOA}
Touvron, H., Lavril, T., Izacard, G., Martinet, X., Lachaux, M.-A., Lacroix, T., Rozi{\`e}re, B., Goyal, N., Hambro, E., Azhar, F., Rodriguez, A., Joulin, A., Grave, E., and Lample, G.
\newblock Llama: Open and efficient foundation language models.
\newblock \emph{ArXiv}, abs/2302.13971, 2023.

\bibitem[Vaswani et~al.(2017)Vaswani, Shazeer, Parmar, and et~al.]{Vaswani2017AttentionIA}
Vaswani, A., Shazeer, N.~M., Parmar, N., and et~al.
\newblock Attention is all you need.
\newblock In \emph{Neural Information Processing Systems}, 2017.

\bibitem[Veli{\v{c}}kovi{\'{c}} et~al.(2018)Veli{\v{c}}kovi{\'{c}}, Cucurull, Casanova, and et~al.]{velickovic2017graphan}
Veli{\v{c}}kovi{\'{c}}, P., Cucurull, G., Casanova, A., and et~al.
\newblock Graph attention networks.
\newblock \emph{International Conference on Learning Representations}, 2018.
\newblock accepted as poster.

\bibitem[Wei et~al.(2022)Wei, Wang, Schuurmans, and et~al.]{Wei2022ChainOT}
Wei, J., Wang, X., Schuurmans, D., and et~al.
\newblock Chain-of-thought prompting elicits reasoning in large language models.
\newblock \emph{Advances in neural information processing systems}, 35:\penalty0 24824--24837, 2022.

\bibitem[Wu et~al.(2019)Wu, Souza, Zhang, Fifty, Yu, and Weinberger]{wu2019simplifying}
Wu, F., Souza, A., Zhang, T., Fifty, C., Yu, T., and Weinberger, K.
\newblock Simplifying graph convolutional networks.
\newblock In \emph{International conference on machine learning}, pp.\  6861--6871. PMLR, 2019.

\bibitem[Wu et~al.(2023)Wu, Zhao, Li, Wipf, and Yan]{Wu2023NodeFormerAS}
Wu, Q., Zhao, W., Li, Z., Wipf, D.~P., and Yan, J.
\newblock Nodeformer: A scalable graph structure learning transformer for node classification.
\newblock \emph{ArXiv}, abs/2306.08385, 2023.
\newblock URL \url{https://api.semanticscholar.org/CorpusID:258509408}.

\bibitem[Xia et~al.(2024)Xia, Zhang, Ye, and et~al.]{Xia2024ChartXC}
Xia, R., Zhang, B., Ye, H., and et~al.
\newblock Chartx \& chartvlm: A versatile benchmark and foundation model for complicated chart reasoning.
\newblock \emph{CoRR}, 2024.

\bibitem[Yang et~al.(2022)Yang, Wu, and Yan]{yang2022geometric}
Yang, C., Wu, Q., and Yan, J.
\newblock Geometric knowledge distillation: Topology compression for graph neural networks.
\newblock \emph{Advances in Neural Information Processing Systems}, 35:\penalty0 29761--29775, 2022.

\bibitem[Zhang et~al.(2022)Zhang, Liu, Sun, and Shah]{zhang2021graph}
Zhang, S., Liu, Y., Sun, Y., and Shah, N.
\newblock Graph-less neural networks: Teaching old mlps new tricks via distillation.
\newblock In \emph{International Conference on Learning Representations}, 2022.

\bibitem[Zhang et~al.(2018)Zhang, Dai, Kozareva, Smola, and Song]{Zhang2017VariationalRF}
Zhang, Y., Dai, H., Kozareva, Z., Smola, A., and Song, L.
\newblock Variational reasoning for question answering with knowledge graph.
\newblock In \emph{Proceedings of the AAAI conference on artificial intelligence}, volume~32, 2018.

\bibitem[Zhu et~al.(2020)Zhu, Yan, Zhao, and et~al.]{Zhu2020BeyondHI}
Zhu, J., Yan, Y., Zhao, L., and et~al.
\newblock Beyond homophily in graph neural networks: Current limitations and effective designs.
\newblock \emph{Advances in neural information processing systems}, 33:\penalty0 7793--7804, 2020.

\bibitem[Zhu et~al.(2023)Zhu, Cong, Zhang, and et~al.]{zhu2023wingnn}
Zhu, Y., Cong, F., Zhang, D., and et~al.
\newblock Wingnn: Dynamic graph neural networks with random gradient aggregation window.
\newblock In \emph{Proceedings of the 29th ACM SIGKDD Conference on Knowledge Discovery and Data Mining}, pp.\  3650--3662, 2023.

\end{thebibliography}
\bibliographystyle{icml2025}

\newpage
\appendix
\onecolumn
\section{Analyzing Order Sensitivity in LLMs and LLMs for Graph Tasks}
\label{Appendix_exp_order}
\begin{figure*}[ht]
\centering
\includegraphics[width=0.8\textwidth]{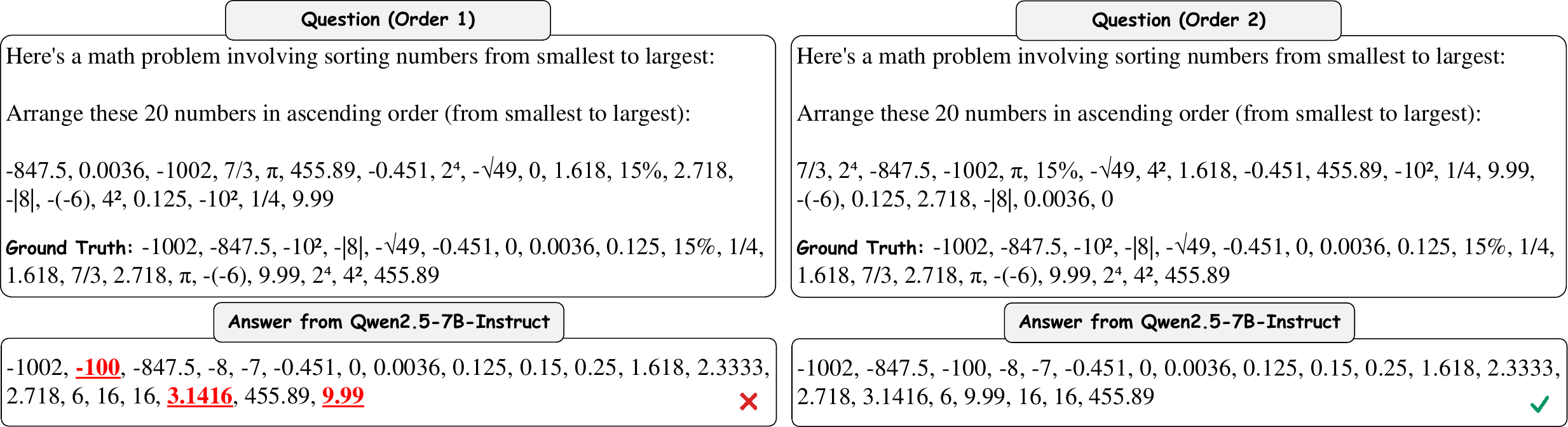} 
\caption{Examples of Qwen2.5-7B-Instruct's responses to the same question with different sequential orderings.}
\label{pic: exporderllm}
\end{figure*}

Many studies find that LLMs are highly sensitive to prompt order in zero-shot and few-shot settings~\cite{Jiang2019HowCW,Lu2021FantasticallyOP,tan2024order}. As shown in Figure~\ref{pic: exporderllm}, taking a sorting problem as an example, Qwen2.5-7B-Instruct~\cite{qwen2_5} produces different results for the same question when presented in two different orders. Counter to intuition, one order leads to the correct answer while the other leads to an incorrect answer. Experiments by Lu et al.~\cite{Lu2021FantasticallyOP} further demonstrate that there is no universally optimal order across different LLMs or tasks.

\begin{figure*}[ht]
\centering
\includegraphics[width=0.95\textwidth]{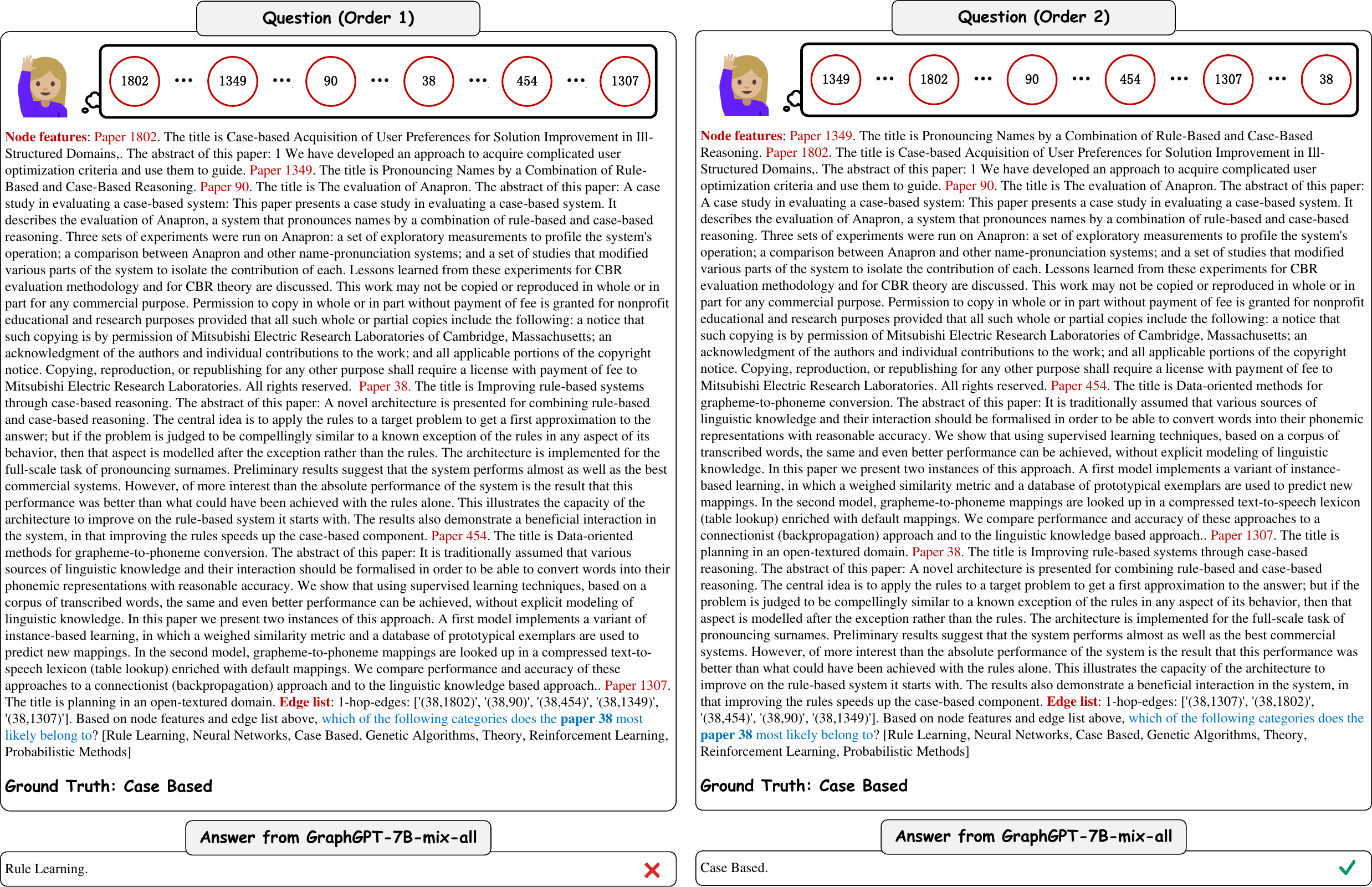} 
\caption{Examples of GraphGPT's responses to node classification tasks where the same graph is described in natural language using different sequential orderings.}
\label{pic: exporder}
\end{figure*}

Similar to the order sensitivity studies on LLMs, we find that Graph LLMs are also order-sensitive. As shown in Figure~\ref{pic: exporder}, taking node classification on the Cora dataset as an example, when we describe the same citation graph using two different natural language orders (i.e., changing the order of elements in Node features and Edge list), GraphGPT~\cite{Tang2023GraphGPTGI}, which is trained on large-scale Text-Attributed Graph data, produces different results for the two orders. One order leads to the correct answer while the other leads to an incorrect answer. This indicates that there exist relatively better natural language description orders for graphs that enable LLMs or Graph LLMs to perform better on graph-related tasks.

\section{Experimental Setup}
\label{Appendix_Experimental_Setup}
We set the learning rate, epoch, batch size, and maximum length for fine-tuning all models to 5e-5, 3, 8, and 1024 respectively. Each experiment is repeated 3 times, with means and standard deviations reported. All experiments use an Intel(R) Xeon(R) Silver 4316 processor as CPU and a single 80G Nvidia A100 GPU. The system memory is 256GB, with Ubuntu 22.03.3 as the operating system, CUDA version 12.4, Python version 3.10.4, and torch version 2.0.1. For Graph CoT distillation in Section~\ref{section:gcot}, we use GPT-4o to generate Graph CoT format answers with a temperature of 0.9 and maximum output tokens of 512.

We employ LoRA (Low-Rank Adaptation) for fine-tuning. The training dataset is preprocessed using 16 workers with a maximum sequence length of 1024 tokens. The LoRA hyperparameters are set as follows: rank = 8, alpha = 16, and dropout = 0, targeting all model layers. For optimization, we use the AdamW optimizer with a learning rate of 5e-5 and cosine learning rate scheduling. We employ mixed-precision training using bfloat16 format. The batch size is set to 2 with a gradient accumulation of 8 steps. Gradient clipping is applied with a maximum norm of 1.0. The model checkpoints are saved every 100 steps, with loss logging occurring every 5 steps. 

For DPO training, we use identical infrastructure settings but configure preference learning with beta = 0.1 and sigmoid-based preference loss. The preprocessing is handled by 16 workers with a maximum sequence length of 2048 tokens. The LoRA parameters remain the same (rank = 8, targeting all layers), while the learning rate is reduced to 5e-6 with cosine scheduling and 10\% warmup ratio. Training uses bfloat16 format over 3 epochs, with batch size = 1 and gradient accumulation steps = 8. Checkpoints are saved every 500 steps, and loss logging occurs every 10 steps.

\section{Additional Results}
\label{Appendix_Additional_Result}

\subsection{Complete Experimental Results}
\label{Appendix_Full_Results}
We also compare with the following baselines, including ChebNet~\cite{defferrard2016convolutional}, SGC~\cite{wu2019simplifying}, and LW-GNN~\cite{pmlr-v198-dai22b}. The results are shown in Table~\ref{tab:full_nodecls} and Table~\ref{tab:full_gqa}.

\begin{table*}[t]
\centering
\small
\caption{Performance comparison (accuracy) on TAG node classification tasks under supervised and zero-shot settings.}
\label{tab:full_nodecls}
\resizebox{\textwidth}{!}{
\begin{tabular}{cccccccc}
\toprule
\multirow{3}{*}{}          & Dataset                & Citeseer           & Cora               & Pubmed             & Cornell            & Texas              & Wisconsin           \\
                           & Edge Hom.              & 0.78               & 0.81               & 0.80               & 0.26               & 0.25               & 0.33                \\
                           & Training Method        & SFT                & 0-shot             & 0-shot             & SFT                & 0-shot             & 0-shot              \\
\midrule
\multirow{7}{*}{\centering GNN} & GCN               & 70.7$\pm$0.4         & 13.9$\pm$3.2         & 26.3$\pm$2.8         & 47.4$\pm$3.9         & 8.9$\pm$7.7          & 21.2$\pm$21.4         \\
                           & GAT                    & 71.2$\pm$0.8         & 13.4$\pm$5.6         & 27.5$\pm$3.3         & 50.5$\pm$2.7         & 37.6$\pm$4.9         & 22.9$\pm$19.2         \\
                           & GraphSage & 70.9$\pm$0.6 & 24.2$\pm$14.1 & 25.8$\pm$3.0 & 48.9$\pm$3.2 & 29.5$\pm$6.8 & 23.5$\pm$20.1 \\
& SGC & 69.8$\pm$0.7 & 13.7$\pm$3.8 & 17.1$\pm$9.9 & 50.8$\pm$3.5 & 9.2$\pm$7.1 & 10.8$\pm$9.8 \\
& ChebNet & 70.3$\pm$0.9 & 23.6$\pm$13.9 & 26.9$\pm$2.7 & 48.9$\pm$3.3 & 9.8$\pm$7.0 & 21.8$\pm$19.5 \\
& LW-GNN & 71.2$\pm$0.7 & 14.0$\pm$3.8 & 35.5$\pm$12.9 & 76.2$\pm$1.0 & 28.9$\pm$6.5 & 24.1$\pm$19.8 \\
& NodeFormer & 70.5$\pm$0.8 & 14.5$\pm$4.0 & 26.1$\pm$3.1 & 75.5$\pm$1.4 & 30.1$\pm$6.6 & 22.8$\pm$20.3 \\
& GKD & 72.0$\pm$0.5 & 14.1$\pm$4.0 & 24.5$\pm$3.2 & 48.2$\pm$3.1 & 9.7$\pm$7.2 & 12.3$\pm$10.0 \\
& GLNN & 73.1$\pm$0.3 & 13.8$\pm$3.7 & 21.0$\pm$2.8 & 51.7$\pm$3.4 & 19.4$\pm$7.4 & 21.6$\pm$19.7 \\
& AdaGMLP & 72.8$\pm$0.4 & 14.1$\pm$3.5 & 11.5$\pm$7.9 & 71.2$\pm$1.3 & 20.1$\pm$7.2 & 22.0$\pm$19.5 \\
\midrule
\multirow{5}{*}{\centering Graph LLM} & GraphWiz    & 74.9$\pm$0.7         & 0.1$\pm$0.9          & 1.5$\pm$1.1          & 50.0$\pm$0.8         & 48.6$\pm$1.2         & 60.6$\pm$0.6         \\
                           & GraphGPT               & 53.2$\pm$1.3         & 9.1$\pm$0.5          & 70.1$\pm$1.4         & 49.8$\pm$0.7         & 52.3$\pm$0.9         & 60.0$\pm$1.1         \\
& GPT-Adjacency & 17.8$\pm$0.5 & 64.2$\pm$0.7 & 20.1$\pm$0.6 & 77.8$\pm$0.1 & 72.9$\pm$0.1 & 79.1$\pm$0.2 \\
& GPT-Incident & 18.6$\pm$0.4 & 65.4$\pm$0.3 & 20.2$\pm$0.7 & 78.2$\pm$0.0 & 73.1$\pm$0.1 & 80.2$\pm$0.0 \\
& GPT-Expert & 18.5$\pm$0.2 & 65.9$\pm$0.3 & 20.8$\pm$0.8 & 78.1$\pm$0.0 & 73.2$\pm$0.1 & 79.9$\pm$0.1 \\
\midrule
\multirow{4}{*}{\centering Qwen 2.5} & 1stage      & 38.4$\pm$0.8         & 36.8$\pm$1.2         & 20.2$\pm$0.6         & 71.9$\pm$1.5         & 64.4$\pm$0.4         & 79.1$\pm$0.9         \\
                           & \textbf{GraphSOS-2stage }        & 64.2$\pm$1.1         & 64.2$\pm$0.7         & 70.8$\pm$1.3         & 75.7$\pm$0.5         & 75.8$\pm$1.4         & 82.1$\pm$0.8         \\
                           & \textbf{GraphSOS-2stage-SSM}     & 65.3$\pm$0.6         & 65.4$\pm$1.4         & 72.3$\pm$0.9         & 77.3$\pm$1.2         & 76.9$\pm$0.3         & 83.5$\pm$1.0         \\
                           & \textbf{GraphSOS-2stage-SSM-OSM} & 69.7$\pm$0.8 & 66.3$\pm$0.6 & 73.9$\pm$1.2 & \textbf{80.1$\pm$0.6} & \textbf{78.6$\pm$0.9} & 84.9$\pm$0.5 \\
\midrule
\multirow{4}{*}{\centering LLaMA 3} & 1stage       & 74.5$\pm$1.2         & 9.7$\pm$0.8          & 7.6$\pm$0.5          & 76.5$\pm$1.3         & 68.5$\pm$0.7         & 79.5$\pm$1.4         \\
                           & \textbf{GraphSOS-2stage}         & 74.9$\pm$0.9         & 67.3$\pm$1.5         & 75.9$\pm$0.4         & 76.5$\pm$0.6         & 72.6$\pm$1.0         & 81.8$\pm$1.2         \\
                           & \textbf{GraphSOS-2stage-SSM}     & 75.3$\pm$0.3         & 68.5$\pm$1.1         & 76.1$\pm$1.4         & 78.9$\pm$0.9         & 74.3$\pm$0.5         & 83.0$\pm$0.8         \\
                           & \textbf{GraphSOS-2stage-SSM-OSM} & \textbf{77.0$\pm$0.5} & \textbf{70.5$\pm$0.8} & \textbf{77.6$\pm$0.7} & 79.5$\pm$0.9 & 76.5$\pm$0.7 & \textbf{85.2$\pm$0.6} \\
\bottomrule
\end{tabular}
}

\end{table*}

\begin{table*}[t]
\centering
\caption{Performance comparison (accuracy) on graph QA tasks under supervised and zero-shot settings.}
\label{tab:full_gqa}
\resizebox{\textwidth}{!}{
\begin{tabular}{cccccccccccc}
\toprule
                           & Dataset            & MetaQA    & cycle     & connect   & bipartite & topology  & shortest  & triangle  & flow      & hamilton   & subgraph   \\
                           & Training Method    & SFT       & SFT       & SFT       & SFT       & SFT       & SFT       & SFT       & SFT       & SFT/0-shot & SFT/0-shot \\
\midrule
\multirow{7}{*}{GNN}       & GCN                & -         & 82.5$\pm$1.3 & 73.0$\pm$0.8 & 81.3$\pm$1.1 & -         & 5.3$\pm$0.7  & 7.0$\pm$1.4  & 10.3$\pm$0.9 & -          & 62.0$\pm$1.2  \\
                           & GAT                & -         & 84.5$\pm$0.6 & 79.8$\pm$1.4 & 83.8$\pm$0.5 & -         & 7.3$\pm$1.2  & 7.3$\pm$0.8  & 11.8$\pm$1.5 & -          & 64.5$\pm$0.4  \\
                           & GraphSAGE & - & 83.8$\pm$1.1 & 78.5$\pm$1.2 & 82.9$\pm$0.8 & - & 7.0$\pm$0.9 & 7.5$\pm$1.1 & 11.2$\pm$1.2 & - & 63.8$\pm$0.8 \\
& SGC & - & 78.2$\pm$0.9 & 78.2$\pm$1.0 & 79.5$\pm$1.3 & - & 6.8$\pm$0.8 & 7.0$\pm$1.0 & 11.5$\pm$1.1 & - & 63.2$\pm$1.0 \\
& ChebNet & - & 80.0$\pm$1.2 & 79.0$\pm$0.9 & 80.2$\pm$0.7 & - & 7.1$\pm$1.1 & 7.4$\pm$0.9 & 11.0$\pm$1.3 & - & 56.0$\pm$0.7 \\
& LW-GNN & - & 79.8$\pm$1.0 & 78.2$\pm$1.1 & 83.0$\pm$0.9 & - & 7.2$\pm$0.8 & 6.3$\pm$1.0 & 12.5$\pm$1.1 & - & 61.5$\pm$0.9 \\
& NodeFormer & - & 83.5$\pm$0.9 & 79.0$\pm$1.0 & 82.7$\pm$1.0 & - & 6.8$\pm$1.0 & 8.7$\pm$0.9 & 13.0$\pm$1.3 & - & 58.2$\pm$0.7 \\
& GKD & - & 84.2$\pm$0.8 & 79.5$\pm$1.1 & 83.1$\pm$0.5 & - & 7.2$\pm$0.8 & 7.2$\pm$1.2 & 11.6$\pm$1.0 & - & 64.2$\pm$0.9 \\
& GLNN & - & 84.7$\pm$1.0 & 80.8$\pm$0.7 & \textbf{83.9$\pm$0.4} & - & 8.2$\pm$0.3 & 7.1$\pm$0.8 & 11.3$\pm$1.4 & - & 63.5$\pm$1.1 \\
& AdaGMLP & - & 84.5$\pm$0.9 & 80.5$\pm$0.8 & 83.6$\pm$0.5 & - & 8.0$\pm$0.4 & 7.3$\pm$0.9 & 11.5$\pm$1.3 & - & 63.8$\pm$1.0 \\
\midrule
\multirow{5}{*}{Graph LLM} & GraphWiz           & 35.3$\pm$1.9  & 70.0$\pm$1.1 & 89.8$\pm$0.3 & 73.3$\pm$1.4 & 16.3$\pm$0.7 & 12.8$\pm$1.0 & 24.0$\pm$0.6 & 28.3$\pm$1.3 & 39.0$\pm$0.8  & 70.3$\pm$1.5  \\
                           & GraphGPT           & 32.9$\pm$1.2 & 72.8$\pm$0.4 & 83.5$\pm$1.5 & 66.8$\pm$0.7 & 0.0$\pm$0.0  & 9.3$\pm$0.5  & 23.3$\pm$1.3 & 13.3$\pm$0.8 & 31.8$\pm$1.4  & 59.8$\pm$0.6  \\
& GPT-Adjacency & 86.9$\pm$1.1 & 81.2$\pm$0.9 & 89.5$\pm$1.0 & 77.8$\pm$0.8 & 70.1$\pm$0.7 & 24.2$\pm$0.7 & 34.8$\pm$1.2 & 36.2$\pm$0.9 & 41.5$\pm$1.1 & 65.1$\pm$0.7 \\
& GPT-Incident & 86.8$\pm$1.1 & 84.9$\pm$0.8 & 90.3$\pm$0.9 & 79.1$\pm$0.7 & \textbf{72.3$\pm$0.8} & 14.5$\pm$0.9 & 25.7$\pm$0.8 & 37.1$\pm$1.1 & 41.2$\pm$0.9 & 66.8$\pm$1.0 \\
& GPT-Expert & 86.9$\pm$1.9 & 81.8$\pm$1.0 & 89.8$\pm$1.1 & 78.2$\pm$0.9 & 70.0$\pm$0.7 & 24.9$\pm$0.6 & 35.1$\pm$1.0 & 36.5$\pm$0.8 & 41.1$\pm$1.2 & 66.8$\pm$0.8 \\
\midrule
\multirow{3}{*}{Qwen 2.5}  & 1stage             & 40.8$\pm$3.5 & 79.3$\pm$1.4 & 88.0$\pm$0.7 & 73.3$\pm$1.2 & 0.0$\pm$0.0  & 10.3$\pm$1.5 & 24.0$\pm$0.9 & 27.3$\pm$0.4 & 31.8$\pm$1.1  & 59.8$\pm$0.8  \\
                           & \textbf{GraphSOS-2stage}     & 80.3$\pm$6.8 & 79.5$\pm$0.6 & 88.0$\pm$1.1 & 74.8$\pm$0.9 & 15.8$\pm$1.4 & 16.8$\pm$0.7 & 21.8$\pm$1.2 & 20.0$\pm$1.5 & 32.8$\pm$0.5  & 68.8$\pm$1.0  \\
                           & \textbf{GraphSOS-2stage-OSM} & 86.9$\pm$3.5 & 80.4$\pm$1.2 & 89.3$\pm$0.6 & 77.1$\pm$0.9 & 16.7$\pm$0.8 & 16.9$\pm$1.1 & 24.2$\pm$0.9 & 26.8$\pm$1.2 & 36.3$\pm$1.2  & 68.9$\pm$0.5  \\
\midrule
\multirow{3}{*}{LLaMA 3}   & 1stage             & 46.4$\pm$3.4 & 83.5$\pm$0.7 & 90.0$\pm$1.2 & 78.5$\pm$0.5 & 0.0$\pm$0.0  & 14.8$\pm$0.8 & 35.3$\pm$1.1 & 25.8$\pm$0.6 & 31.8$\pm$1.3  & 62.5$\pm$0.9  \\
                           & \textbf{GraphSOS-2stage }    & 83.3$\pm$1.1 & 89.8$\pm$1.3 & 92.8$\pm$0.8 & 79.3$\pm$1.4 & 17.3$\pm$0.4 & 24.5$\pm$1.1 & 37.5$\pm$0.5 & 38.5$\pm$1.2 & 41.0$\pm$0.7  & 69.5$\pm$1.5  \\
                           & \textbf{GraphSOS-2stage-OSM} & \textbf{89.6$\pm$1.0} & \textbf{92.7$\pm$0.7} & \textbf{93.4$\pm$1.2} & 80.2$\pm$0.9 & 18.8$\pm$0.8 & \textbf{26.3$\pm$0.5} & \textbf{41.2$\pm$1.3} & \textbf{40.7$\pm$1.0} & \textbf{42.7$\pm$0.9}  & \textbf{72.9$\pm$0.8} \\
\bottomrule
\end{tabular}
}

\end{table*}

\subsection{Module Ablation Results}
\label{Appendix_Ablation_ Results}
\begin{figure}[ht]
\centering
\begin{minipage}{0.48\textwidth}
    \centering
    \includegraphics[width=\textwidth]{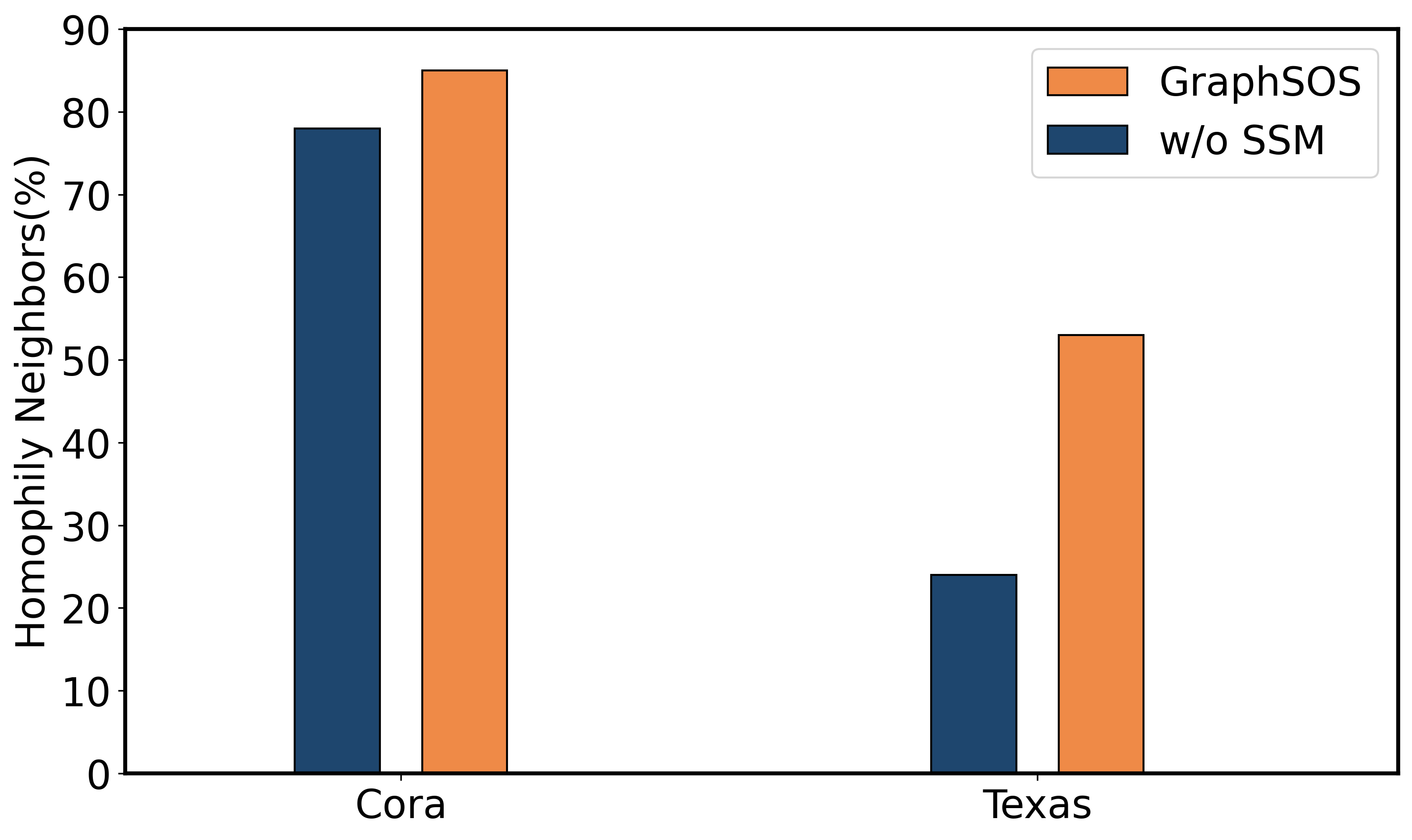}
    \caption{Proportion of same-class neighbors in random sampling and SSM sampling.}
    \label{pic: ablation_ssm}
\end{minipage}
\hfill
\begin{minipage}{0.48\textwidth}
    \centering
    \includegraphics[width=\textwidth]{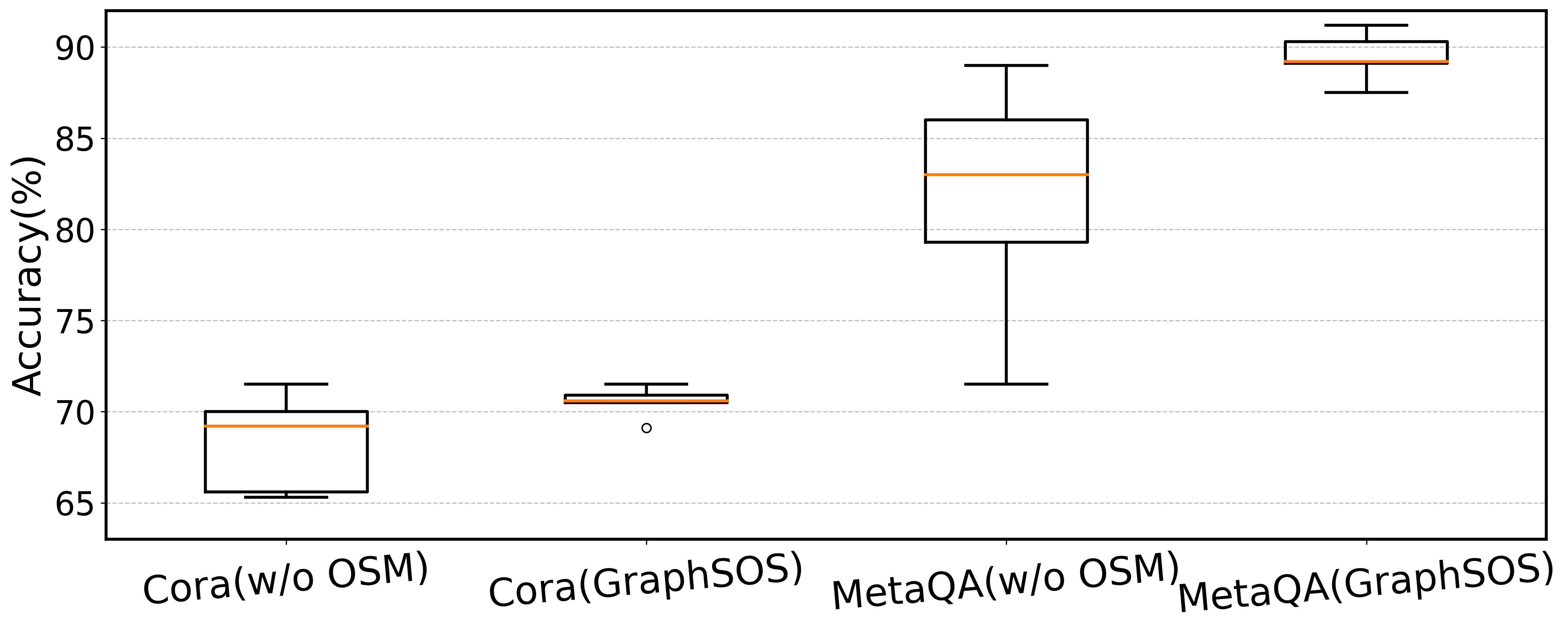}
    \caption{Performance fluctuation comparison between random order and OSM-selected order.}
    \label{pic: ablation_osm}
\end{minipage}
\end{figure}


\subsection{Parameter Sensitivity (RQ3)}
\label{Appendix_RQ3}
\begin{figure}[ht]
\centering
\includegraphics[width=0.48\textwidth]{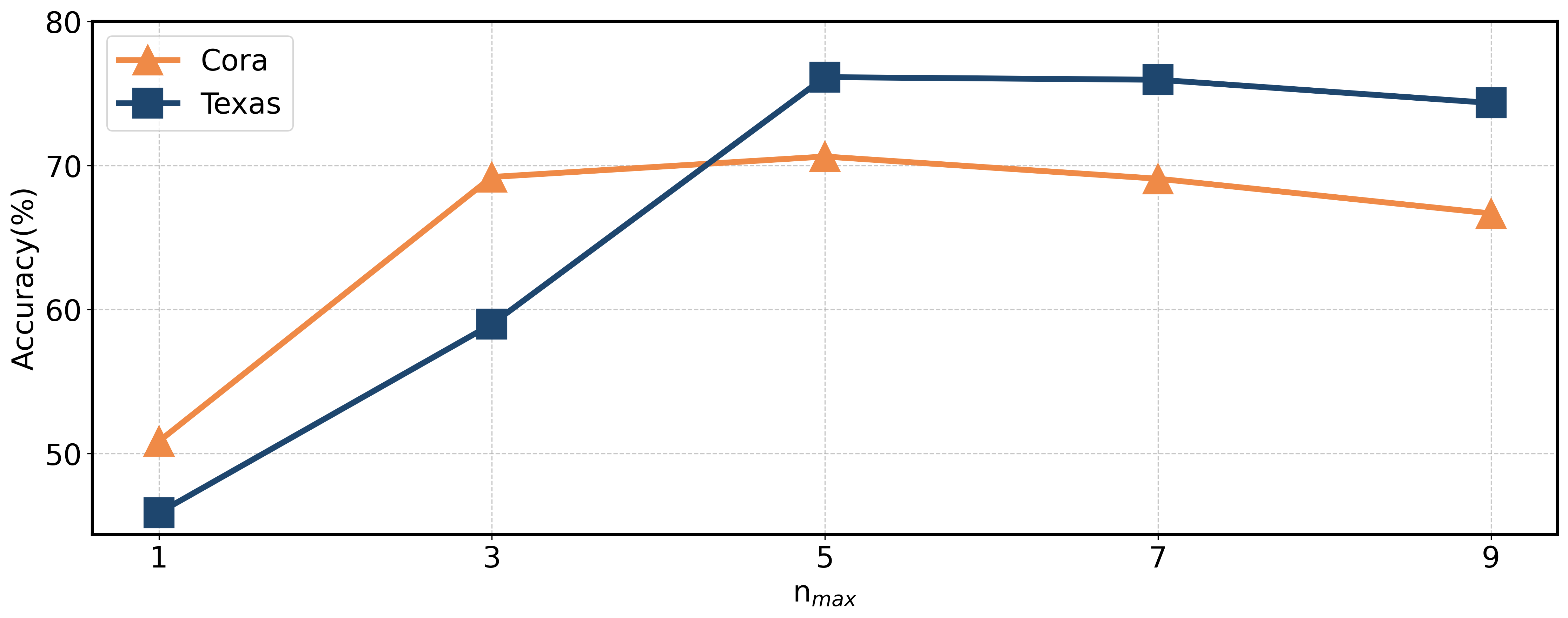} 
\caption{Performance of GraphSOS with different numbers of sampled neighbors $n_\text{max}$.}
\label{pic: ablation_nmax}
\end{figure}

We analyze the impact of two hyperparameters on GraphSOS: the number of sampled neighbors $n_\text{max}$ in SSM and the number of order candidates $m$ in OSM. Figure~\ref{pic: ablation_nmax} shows the performance of GraphSOS with LLaMA 3-8B as the base model under different settings of sampled neighbors $n_\text{max}$. Results indicate that both too high and too low values of $n_\text{max}$ affect model performance. A low $n_\text{max}$ leads to limited graph structural information for LLM, restricting model reasoning; a high $n_\text{max}$ results in overly long context input to LLM, making reasoning difficult~\cite{An2024MakeYL}. Table~\ref{tab: ablation_m} shows the performance of LLaMA 3-8B-based GraphSOS under different order candidate numbers $m$. The results indicate that the model performance improves steadily as $m$ increases. This suggests that appropriately increasing $m$ can enhance performance by including more candidate order samples. However, increasing the order leads to growth in inference time overhead. To balance performance and time overhead, we only choose to set $m=10$. Nevertheless, we point out that increasing $m$ brings performance improvements.

\begin{table}[ht]
\centering
\footnotesize
\caption{Accuracy and inference time of baselines and GraphSOS with different numbers of order candidates $m$ on MetaQA.}
\label{tab: ablation_m}
\begin{tabular}{@{}c|c|c@{}}
\toprule
\textbf{Models} & \textbf{Acc(\%)} & \textbf{time(s)} \\ \midrule
GraphWiz & 35.3$\pm$1.9 & \textbf{0.43} \\
GraphGPT & 32.9$\pm$1.2 & 0.51 \\
LLaMa 3 & 46.4$\pm$3.4 & 0.65 \\ \midrule
\textbf{GraphSoS ($m=5$)} & 84.5$\pm$0.5 & 0.81 \\
\textbf{GraphSoS ($m=10$)} & 89.6$\pm$1.0 & 1.02 \\
\textbf{GraphSoS ($m=15$)} & 89.8$\pm$1.2 & 1.19 \\
\textbf{GraphSoS ($m=20$)} & \textbf{90.2$\pm$0.9} & 1.37 \\ \bottomrule
\end{tabular}
\end{table}

\section{Future Work}
In this paper, when discussing graph structures, we focus on homophily and heterophily. However, other structural properties (such as degree, connectivity, symmetry) are worth exploring. We demonstrate that with properly constructed training data, subgraphs with any graph structure that is more beneficial for tasks can be sampled by training GraphSOS's SSM, rather than only homophily and heterophily. Additionally, ensuring accurate Graph CoT steps is crucial for graph reasoning. We encourage introducing more powerful models to generate Graph CoT data or using manually constructed Graph CoT data, and emphasize the effectiveness of constructing larger training datasets to improve model performance. Moreover, the Subgraph Sampling Module constructed in this paper lacks explicit structure-aware components. We encourage new methods to incorporate structure-awareness capabilities to discover, evaluate, and preserve critical paths or paths that play important connecting roles in the topological structure.



\end{document}